\documentclass[runningheads]{llncs}

\usepackage{color}
\usepackage[width=122mm,left=12mm,paperwidth=146mm,height=193mm,top=12mm,paperheight=217mm]{geometry}

\usepackage{times}
\usepackage{epsfig}
\usepackage{graphicx}
\usepackage{amsmath}
\usepackage{amssymb}
\usepackage{bbm}
\usepackage{dsfont}
\usepackage{subfig}
\usepackage{tikz}
\usepackage{multirow}

\usepackage[pagebackref=true,breaklinks=true,letterpaper=true,colorlinks,bookmarks=false]{hyperref}
\usepackage{algorithm,algcompatible}
\algnewcommand\INPUT{\item[\textbf{Input:}]}%
\algnewcommand\OUTPUT{\item[\textbf{Output:}]}%

\begin{document}

\def\argmax{\operatornamewithlimits{argmax}}
\def\argmin{\operatornamewithlimits{argmin}}
\def\Tr{Tr}
\algdef{SE}[DOWHILE]{Do}{doWhile}{\algorithmicdo}[1]{\algorithmicwhile\ #1}%

\newcommand{\etal}{~\emph{et~al.}}


\newcounter{nbdrafts}
\setcounter{nbdrafts}{0}
\makeatletter
\newcommand{\checknbdrafts}{
\ifnum \thenbdrafts > 0
\@latex@warning@no@line{**********************************************************************}
\@latex@warning@no@line{* The document contains \thenbdrafts \space draft note(s)}
\@latex@warning@no@line{**********************************************************************}
\fi}
\newcommand{\draft}[1]{\addtocounter{nbdrafts}{1}{\color{red} #1}}
\makeatother

\newcommand{\vstretch}{\vspace*{\stretch{1}}}
\newcommand{\hstretch}{\hspace*{\stretch{1}}}

\pagestyle{headings}
\mainmatter
\def\ECCV16SubNumber{1140}  
\title{Scalable Metric Learning via Weighted Approximate Rank Component Analysis}


\author{Cijo  Jose\qquad Fran\c{c}ois Fleuret}
\institute{Idiap Research Institute\\
\'{E}cole Polytechnique F\'{e}d\'{e}rale de Lausanne\\
{\tt\small \{firstname.lastname\}@idiap.ch}
}

\maketitle



\begin{abstract}

We are interested in the large-scale learning of Mahalanobis
distances, with a particular focus on person re-identification.

We propose a metric learning formulation called Weighted Approximate
Rank Component Analysis (WARCA). WARCA optimizes the precision at top
ranks by combining the WARP loss with a regularizer that favors
orthonormal linear mappings, and avoids rank-deficient
embeddings. Using this new regularizer allows us to adapt the
large-scale WSABIE procedure and to leverage the Adam stochastic
optimization algorithm, which results in an algorithm that scales
gracefully to very large data-sets. Also, we derive a kernelized
version which allows to take advantage of state-of-the-art features
for re-identification when data-set size permits kernel computation.

Benchmarks on recent and standard re-identification data-sets show
that our method beats existing state-of-the-art techniques both in
term of accuracy and speed. We also provide experimental analysis to
shade lights on the properties of the regularizer we use, and how it
improves performance.

\end{abstract}









\section{Introduction} \label{introduction}

Metric learning methods aim at learning a parametrized distance function from a labeled set of samples, so that under the learned distance, samples with the same labels are nearby and samples with different labels are far apart~\cite{Weinberger2009}. Many fundamental questions in computer vision such as ``How to compare two images? and for what information?'' boil down to this problem. Among them, person re-identification is the problem of recognizing  individuals at different physical locations and times, on images captured by different devices.

It is a challenging problem which recently received a lot of attention because of its importance in various application domains such as video surveillance, biometrics and behavior analysis ~\cite{Gong2014}.

The performance of person re-identification systems relies mainly on the image feature representation and the distance measure used to compare them. Hence the research in the field has focused either on designing features~\cite{Dong2011,Zhao2014,LiaoXQDA2015} or on learning a distance function from a labeled set of images~\cite{Mignon2012,Kostinger2012,Li2013,Xiong2014,LiaoXQDA2015,LiaoMLAPG2015}.


It is difficult to analytically design features that are invariant to the various non-linear transformations that an image undergoes such as illumination, viewpoint, pose changes and occlusion.  Furthermore, even if such features were provided, the standard Euclidean metric would not be adequate as it does not take into account dependencies on the feature representation. This motivates the use of metric learning for person re-identification.


Re-identification models are commonly evaluated by the cumulative match characteristic (CMC) curve. This measure indicates how the matching performance of the algorithm  improves as the number of returned image increases. Given a matching algorithm and a labeled test set, each image is compared against all the others, and the position of the first correct match is recorded. The CMC curve indicates for each rank the fraction of test samples which had that rank or  better. A perfect CMC curve would reach the value $1$ for rank $\#1$, that is the best match is always of the correct identity.

In this paper we are interested in learning a Mahalanobis distance by minimizing a weighted rank loss such that the precision at the top rank positions of the CMC curve is maximized. When learning the metric, we directly learn the low-rank projection matrix instead of the PSD matrix because of the computational efficiency and the scalability to high dimensional datasets (see~\S~\ref{sec:problem-formulation}). But naively learning the low-rank projection matrix suffers from the problem of matrix rank degeneration and non-isolated minima~\cite{Lim2014}. We address this problem by using a simple regularizer which approximately enforces the orthonormality of the learned matrix very efficiently (see~\S~\ref{sec:appr-orth-regul}). We extend the WARP loss~\cite{Usunier2009,Weston2011,Lim2014} and combine it with our approximate orthonormal regularizer to derive a metric learning algorithm which approximately minimizes a weighted rank loss efficiently using stochastic gradient descent (see~\S~\ref{sec:max-marg-reform}).

We extend our model to kernel space to handle distance measures which are more natural for the features we are dealing with (see~\S~\ref{sec:kernelization}). We also show that in kernel space SGD can be carried out more efficiently by using preconditioning~\cite{Chapelle2007,Mignon2012}.


We validate our approach on nine challenging person re-identification datasets: {Market-1501}~\cite{Zheng2015}, {CUHK03}~\cite{Li2014}, {OpeReid}~\cite{LiaoMHL14}, {CUHK01}~\cite{Wei2012}, {VIPeR}~\cite{Gray2008}, {CAVIAR}~\cite{Dong2011}, {3DPeS}~\cite{Baltieri2011}, {iLIDS}~\cite{Zheng2009} and {PRI450s}~\cite{Roth2014}, where we  outperform other metric learning methods proposed in the literature, both in speed and accuracy.

\section{Related Works} \label{related_works}

Metric learning is a well studied research problem~\cite{Yang2006}. Most of the existing  approaches have been developed in the context of the Mahalanobis distance learning paradigm \cite{Xing2002,Weinberger2009,Davis2007,Mignon2012,Kostinger2012}.  This consists in learning distances of the form:

\vspace*{-1em}

\begin{equation}
\mathcal{D}_{M}^2(x_i, x_j) = (x_i-x_j)^TM(x_i-x_j),
\end{equation}
where $M$ is a positive semi-definite matrix.
Based on the way the problem is formulated the algorithms for learning such distances involve either
optimization in the space of positive semi-definite (PSD) matrices, or learning the projection matrix $W$, in which case $M = W^TW$.

Large margin nearest neighbors~\cite{Weinberger2009} (LMNN) is a metric learning algorithm designed to maximize the performance of $k$-nearest neighbor classification in a large margin framework. Information theoretic metric learning~\cite{Davis2007} (ITML) exploits the relationship between the Mahalanobis distance and Gaussian distributions to learn a metric by minimizing the KL-divergence with a metric prior. Many researchers have applied LMNN and ITML to re-identification problem with varying degree of success~\cite{Roth2014}.

Pairwise Constrained Component Analysis (PCCA)~\cite{Mignon2012} is a metric learning method that learns the low rank projection matrix $W$ in the kernel space from sparse pairwise constraints.  Xiong\etal~\cite{Xiong2014}  extended PCCA with a  $L_2$ regularization term and showed that it further improves the performance.

K\"ostinger\etal~\cite{Kostinger2012} proposed the KISS (``Keep It Simple and Straight forward'') metric learning abbreviated as KISSME. Their method enjoys very fast training and they show good empirical performance and scaling properties along the number samples. However this method suffers from of the Gaussian assumptions on the model.

Li\etal~\cite{Li2013} consider learning a local thresholding rule for metric learning. This method is computationally expensive to train, even with as few as $100$ dimensions. Their paper discusses solving the problem in dual to decouple the dependency on the dimension but in practice it is solved in primal form with off-the-shelf solvers.

The performance of many kernel-based metric learning-based methods for person re-identification was evaluated in~\cite{Xiong2014}. In particular the authors evaluated  PCCA~\cite{Mignon2012},  variants of kernel Fisher discriminant analysis (KFDA) and reported that the KFDA variants consistently out-perform all other methods. The KFDA variants they investigated were Local Fisher Discriminant Analysis (LFDA) and Marginal Fisher Discriminant Analysis(MFA).

Recently few metric learning algorithms have been proposed for person re-identification. Chen\etal~\cite{Chen2015} attempt to learn a metric in the polynomial feature map exploiting the relationship between Mahalanobis metric and the polynomial features. Ahmed\etal~\cite{Ahmed2015} propose a deep learning model which learns the features as well as the metric jointly. Liao\etal~\cite{LiaoXQDA2015} propose XQDA exploiting the benefits of Fisher discriminant analysis and KISSME to learn a metric. However like FDA and KISSME, XQDA's modeling power is limited because of the Gaussian assumptions on the data. In another work Liao\etal~\cite{LiaoMLAPG2015} apply accelerated proximal gradient descent (APGD) to a Mahalanobis metric under a logistic loss similar to the loss of PCCA~\cite{Mignon2012}. The proximal step is the projection on to the PSD cone. They also use asymmetric sample weighting to exploit the imbalance between the positive and negative pairs. The application of APGD makes this model converge fast compared to existing batch metric learning algorithms but still it suffers from scalability issues because all the pairs are required to take one gradient step and the projection step on to the PSD cone is computationally expensive.

None of the above mentioned techniques explicitly models the objective that we are looking for in person re-identification, that is to optimize a weighted rank measure.
%
%
We show that modeling this in the metric learning objective improves the performance. We address scalability through stochastic gradient descent and our model naturally eliminates the need for asymmetric sample weighting as we use triplet based loss function.

There is an extensive body of work on optimizing ranking measures such as AUC, precision at $k$, $F_{1}$ score, etc. Most of this work focuses on learning a linear decision boundary in the original input space, or in the feature space for ranking a list of items based on the chosen performance measure. A well known such model is the structural SVM~\cite{Tsochantaridis2004}.
In contrast here we are interested in ranking pairs of items by learning a metric. A related work by McFee\etal~\cite{Mcfee2010} studies metric learning with different rank measures in the  structural SVM framework. Wu\etal~\cite{Wu2011} used this framework to do person re-identification by optimizing the mean reciprocal rank criterion. Outside the direct scope of metric learning from a single feature representation, Paisitkriangkrai\etal~\cite{Paisitkriangkrai2015} developed an ensemble algorithm to combine different base metrics in the structural SVM framework which leads to excellent performance for re-identification. Such an approach is complementary to ours, as combining heterogeneous feature representations requires a separate additional level of normalization or the combination with a voting scheme.

We use the WARP loss from WSABIE~\cite{Weston2011}, initially proposed for large-scale image annotation problem, that is a multi-label classification problem. WSABIE simultaneously optimizing the precision in the top rank positions and learn a low dimensional joint embedding for both images and annotations. This work reports excellent empirical results in terms of accuracy, computational efficiency, and memory footprint.

The work that is closely related to us is FRML~\cite{Lim2014} where they learn a Mahalanobis metric by optimizing the WARP loss function with stochastic gradient descent. However there are some key differences with our approach. FRML is a linear method using $L_2$ or LMNN regularizer, and relies on an expensive projection step in the SGD. Beside, this projection requires to keep a record all the gradients in the mini-batch, which results in high memory footprint.  The rationale for the projection step is to accelerate the SGD because directly optimizing  low rank matrix may result in rank deficient matrix and thus results in non-isolated minimizers which might generalizes poorly to unseen samples. We propose a computationally cheap solution to this problem by using a regularizer which approximately enforces the rank of the learned matrix very efficiently.

\begin{table}
\caption{Notation}
\center
\begin{tabular}{ll}
\hline
$N$ \ Number of training samples \\
$D$ \ Dimension of training samples \\
$Q$ \ Number of classes \\
$(x_i, y_i) \in \mathbb{R}^D \times \{ 1, \dots, Q \}$  $i$-th training sample \\
$\mathds{1}_{\text{condition}}$ is equal to 1 if the condition is true, 0 otherwise   \\
$\mathcal{S}$ the pairs of indices of samples of same class \\
$\mathcal{T}_{{y}}$ the indices of samples not of class $y$ \\
$\mathcal{F}_W$ distance function under the linear map $W$ \\
$rank_{i,j}(\mathcal{F}_W)$ for $i$ and $j$ of same class, number of miss-labeled examples closer to $i$ than $j$ is \\
$\mathcal{L}(W)$ the loss we minimize \\
$L(r)$ rank weighting function \\
\hline
\end{tabular}
\label{table_notations}
\end{table}

\section{Weighted Approximate Rank Component Analysis} \label{method}
This section presents our metric learning algorithm, Weighted Approximate Rank Component Analysis (WARCA). Table~\ref{table_notations} summarizes some important notations that we use in the paper.

Let us consider a training set of data point / label pairs:
\begin{equation}
(x_n, y_n) \in \mathbb{R}^D \times \{ 1, \dots, Q \}, \ n = 1, \dots ,N.
\end{equation}
and let $\mathcal{S}$ be the set of pairs of indices of samples of same labels:
\begin{equation}
\mathcal{S} = \left\{ (i, j) \in \{1, \dots, N \}^{2}, \ y_i = y_j \right\}.
\end{equation}
For each label ${y}$ we define the set $\mathcal{T}_{{y}}$ of indices of samples of a class different from $y$:
\begin{equation}
\mathcal{T}_{{y}} = \left\{k  \in \{1, \dots, N \}, \ y_k \neq {y} \right\}.
\end{equation}
In particular, to each $(i, j) \in \mathcal{S}$ corresponds a set $\mathcal{T}_{y_{i}}$.

Let $W$ be a linear transformation that maps the data points from $\mathbb{R}^D$ to $\mathbb{R}^{D'}$, with $D' \leq D$. For the ease of notation, we do not distinguish between matrices and their corresponding linear mappings. The distance function under the linear map $W$ is given by:
\begin{equation}
  \mathcal{F}_{W}(x_i, x_j) = \| W (x_i - x_j ) \|_2.
  \label{equation_0}
\end{equation}

\subsection{Problem Formulation}\label{sec:problem-formulation}

For a pair of points $(i, j)$ of same label $y_i = y_j$, we define a
ranking error function:
\begin{equation}
\forall (i, j) \in \mathcal{S}, \ err(\mathcal{F}_{W}, i, j) = L\left(rank_{i,j}\left( \mathcal{F}_{W} \right)\right)
\end{equation}
where:
\begin{equation}
rank_{i,j}\left( \mathcal{F}_{W} \right)
=
\sum_{y_i \neq y_k} I\left( \mathcal{F}_{W}(x_i, x_k) \leq \mathcal{F}_{W}(x_i, x_j) \right)
\end{equation}
is the number of samples $x_k$ of different labels which are closer to
$x_i$ than $x_j$ is.

Formulating our objective that way, following closely the formalism of
\cite{Weston2011}, shows how training a multi-class predictor shares
similarities with our metric-learning problem. The former aims at
avoiding, for any given sample to have incorrect classes with
responses higher than the correct one, while the latter aims at
avoiding, for any pair of samples $(x_i, x_j)$ of the same label, to
have samples $x_k$ of other classes in between them.

Minimizing directly the rank treats all the rank positions equally, and
usually in many problems including person re-identification we are
interested in maximizing the correct match within the top few rank
positions. This can be achieved by a weighting function ${{L}}(\cdot)$
which penalizes more a drop in the rank at the top positions than at
the bottom positions. In particular we use the rank weighting function
proposed by Usunier\etal~\cite{Usunier2009}, of the form:
\begin{equation}
{{L}}(r) = \sum_{s=1}^{r} \alpha_s, \ \alpha_1 \geq \alpha_2 \geq ... \geq 0.
\label{equation_wr}
\end{equation}
For example, using $\alpha_1=\alpha_2=...=\alpha_m$ will treat all
rank positions equally, and using higher values of $\alpha$s in top
few rank positions will weight top rank positions more. We use the
harmonic weighting, which has such a profile and was also used in
\cite{Weston2011} as it yielded state-of-the-art results on their
application.

Finally, we would like to solve the following optimization problem:
\begin{equation}
  \argmin_{W} \ \frac{1}{|\mathcal{S}|}  \sum_{(i, j) \in \mathcal{S}}  {{L}}\left(rank_{i,j}\left( \mathcal{F}_{W} \right)\right).
  \label{equation_1}
\end{equation}

\subsection{Approximate OrthoNormal (AON) Regularizer}\label{sec:appr-orth-regul}

The optimization problem of Equation~\ref{equation_1} is made more
difficult by training sets of small or medium sizes, which may lead to
a severe over-fitting. Regularizing penalty terms are central in
re-identification for that reason.

The standard way of regularizing a low-rank metric learning objective
function is by using a $L_2$ penalty, such as the Frobenius
norm~\cite{Lim2014}. However, such a regularizer tends to push toward
rank-deficient linear mappings, which we observe in practice (see
\S~\ref{sec:analys-aon-regul}, and in particular
Figure~\ref{ref_cond}).


Lim\etal~\cite{Lim2014} in their FRML algorithm, addresses this problem by using a
Riemannian manifold update step in their SGD algorithm, which is
computationally expensive and induces a high memory footprint. 
We propose an alternative approach that maintains the rank of the
matrix by pushing toward orthonormal matrices. This is achieved by
using as a penalty term the $L_2$ divergence of $W W^T$ from the
identity matrix $\mathbf{I}$ :
\begin{equation}
\|W W^T - \mathbf{I}\|^2. \label{eq:regularizer}
\end{equation}

This orthonormal regularizer can also be seen as a strategy to mimic
the behavior of approaches such as PCA or FDA, which ensure that the
learned linear transformation is orthonormal. For such methods, this
property emerges from the strong Gaussian prior over the data, which
is beneficial on small data-sets but degrades performance on large
ones where it leads to under-fitting. Controlling the orthonormality
of the learned mapping through a regularizer weighted by a
meta-parameter $\lambda$ allows us to adapt it on each data-set
individually through cross-validation.

Finally, with this regularizer the optimization problem of
Equation~\ref{equation_1} becomes:
\begin{equation}
    \argmin_{W} \ \frac{\lambda}{2} \| W W^T - \mathbf{I} \|^2  + \frac{1}{|\mathcal{S}|}  \sum_{(i, j) \in \mathcal{S}}  {{L}}\left(rank_{i,j}\left( \mathcal{F}_{W} \right)\right).
  \label{equation_2}
\end{equation}

\subsection{Max-Margin Reformulation}\label{sec:max-marg-reform}
The metric learning problem on Equation~\ref{equation_2} aims at minimizing the $0$-$1$ loss, which is a difficult optimization problem. Applying the reasoning behind the WARP loss to make it tractable, we upper-bound this loss with the hinge one with margin $\gamma$.  This is equivalent to minimizing the following loss function:
\begin{multline}
\begin{aligned}
\hspace*{-1.2em}\mathcal{L}(W) = \, \, & \frac{\lambda}{2}\|WW^T - \mathbf{I}\|^2 \ +  \frac{1}{|\mathcal{S}|} \sum_{(i, j) \in \mathcal{S}}  \sum_{k \in \mathcal{T}_{{y_{i}}}}
        {{L}}({rank}^\gamma_{i,j}({\cal F}_W))\frac{\left|\gamma + \xi_{{i}{j}{k}} \right|_+}{{rank}^\gamma_{i,j}({\cal F}_W)}, \hspace*{-0.7em}
\label{equation_3}
\end{aligned}
\end{multline}
where:
\begin{equation}
\xi_{{i}{j}{k}} =  \mathcal{F}_{W}(x_{i}, x_{j}) - \mathcal{F}_{W}(x_{i}, x_{k})
\end{equation}
and  ${rank}^\gamma_{i,j}({\cal F}_W)$ is the margin penalized rank:
\begin{equation}
{rank}^\gamma_{i,j}({\cal F}_W) = \sum_{k \in \mathcal{T}_{{y_{i}}}} \mathds{1}_{\gamma + \xi_{{i}{j}{k}} > 0}.
\end{equation}

The loss function in Equation~\ref{equation_3} is the WARP loss~\cite{Usunier2009,Weston2011,Lim2014}. It was shown by Weston\etal~\cite{Weston2011} that the WARP loss can be efficiently solved by using stochastic gradient descent and we follow the same approach:
\begin{enumerate}
\item Sample $(i, j)$ uniformly at random from $\mathcal{S}$.
\item For the selected $(i, j)$ uniformly sample $k$ in $\left\{ k \in \mathcal{T}_{{y_{i}}}: \gamma + \xi_{{i}{j}{k}} > 0 \right\}$, \emph{i.e.} from the set of incorrect matches scored higher than the correct match $x_j$.
\end{enumerate}
The sampled triplet $(i, j, k)$ has a contribution of ${{L}}({rank}^\gamma_{i,j}({\cal F}_W))|\gamma + \xi_{{i}{j}{k}} |_+$
because the probability of drawing a $k$  in step 2 from the violating set is $\frac{1}{{rank}^\gamma_{i,j}({\cal F}_W)}$.

We use the above sampling procedure to solve WARCA efficiently using mini-batch stochastic gradient descent (SGD). We use Adam SGD algorithm~\cite{Kingma2014}, which is found to converge faster empirically compared to vanilla SGD.  

\subsection{Kernelization}\label{sec:kernelization}

Most commonly used features in person re-identification are histogram-based such as LBP, SIFT BOW, RGB histograms to name a few. The most natural distance measure for histogram-based features is the $\chi^{2}$ distance. Most of the standard metric learning methods work on the Euclidean distance with PCCA being a notable exception. To plug any arbitrary metric which is suitable for the features, such as $\chi^2$, one has to resort to explicit feature maps that approximates the $\chi^{2}$ metric. However, it blows up the dimension and the computational cost. Another way to deal with this problem is to do metric learning in the kernel space, which is the approach we follow.

Let $W$ be spanned by the samples:
\begin{equation}
W = A X^T = A
\left(
\begin{array}{c}
x_1^T   \\
\dots \\
x_N^T
\end{array}
\right).
\end{equation}
which leads to:
\begin{eqnarray}
\mathcal{F}_{A}(x_i, x_j) & = & \| A X^T(x_i - x_j) \|_2,                          \\
                          & = & \| A ({\kappa}_i - {\kappa}_j) \|_2.
\end{eqnarray}
Where ${\kappa}_i$ is the $i^{th}$ column of the kernel matrix $K = X^TX$.
Then the loss function in Equation~\ref{equation_3} becomes:
\begin{multline}
\hspace*{-1.0em}\mathcal{L}(A)= \frac{\lambda}{2}\|AKA^T-\mathbf{I}\|^2  + \frac{1}{|\mathcal{S}|}\sum_{(i, j) \in \mathcal{S}}  \sum_{k \in \mathcal{T}_{{y_{i}}}} {{L}}({rank}^\gamma_{i,j}({\cal F}_A)) \frac{|\gamma +\xi_{{i}{j}{k}} |_+}{{rank}^\gamma_{i,j}({\cal F}_A)},
\label{equation_4}
\end{multline}
with:
\begin{equation}
\xi_{{i}{j}{k}} =  \mathcal{F}_{A}(x_{i}, x_{j}) - \mathcal{F}_{A}(x_{i}, x_{k}).
\end{equation}

Apart from being able to do non-linear metric learning, kernelized WARCA can be solved efficiently again by using stochastic sub-gradient descent.
If we use the inverse of the kernel matrix as the pre-conditioner of the stochastic sub-gradient, the computation of the update equation, as well the parameter update, can be carried out efficiently. Mignon\etal~\cite{Mignon2012} used the same technique to solve their PCCA, and showed that it converges faster than vanilla gradient descent. We use the same technique to derive an efficient update rule for our kernelized WARCA. A  stochastic sub-gradient of Equation~\ref{equation_4} with the sampling procedure described in the previous section is given as:
\begin{multline}
\nabla \mathcal{L}(A) = 2\lambda(AKA^T-\mathbf{I})AK + 2{{L}}({rank}^\gamma_{i,j}({\cal F}_A))A\mathds{1}_{{\gamma + \xi_{{i}{j}{k}} > 0}} \mathcal{G}_{ijk} \label{equation_7},
\end{multline}
where:
\begin{equation}
\mathcal{G}_{ijk} = \frac{({\kappa}_{i}-{\kappa}_{j})({\kappa}_{i}-{\kappa}_{j})^T}{d_{ij}} - \frac{({\kappa}_{i}-{\kappa}_{k})({\kappa}_{i}-{\kappa}_{k})^T}{d_{ik}},
\end{equation}
and:
\begin{equation}
d_{ij} =  \mathcal{F}_{A}(x_{i}, x_{j}), \ \ d_{ik} =  \mathcal{F}_{A}(x_{i}, x_{k}).
\end{equation}

Multiplying the right hand side of Equation~\ref{equation_7} by $K^{-1}$:
\begin{multline}
\nabla \mathcal{L}(A)K^{-1} = 2\lambda(AKA^T-\mathbf{I})A + 2{{L}}({rank}^\gamma_{i,j}({\cal F}_A))AK\mathds{1}_{{\gamma + \xi_{{i}{j}{k}} > 0}}\mathcal{E}_{ijk} \label{equation_9}.
\end{multline}
with:
\begin{eqnarray} 
\hspace*{-2em}
\mathcal{E}_{ijk} =  K^{-1}\mathcal{G}_{ijk} K^{-1} = \frac{(e_{i}\!-\!e_{j})(e_{i}\!-\!e_{j})^T}{d_{ij}} - \frac{(e_{i}\!-\!e_{k})(e_{i}\!-\!e_{k})^T}{d_{ik}}  \label{equation 10}.
\end{eqnarray}
where $e_l$ is the $l^{th}$ column of the canonical basis that is the vector whose $l^{th}$ component is one and all others are zero.
In the preconditioned stochastic sub-gradient descent we use the updates of the form:
\begin{multline}
A_{t+1} = (\mathbf{I} - 2\lambda \eta(A_{t}KA_{t}^T-\mathbf{I}))A_{t} - 2\eta{{L}}({rank}^\gamma_{i,j}({\cal F}_A))A_{t}K\mathds{1}_{\gamma + \xi_{{i}{j}{k}} > 0}\mathcal{E}_{ijk}.
\end{multline}
Please note that $\mathcal{E}_{ijk}$ is a very sparse matrix with only nine non-zero entries. This makes the update extremely fast. Preconditioning also enjoys faster convergence rates since it exploits second order information through the preconditioning operator, here the inverse of the kernel matrix \cite{Chapelle2007}.

\section{Experiments}

We evaluate our proposed algorithm on nine standard person re-identification datasets. We first describe the datasets and baseline algorithms and then present our results. The source-code of our experimental framework, including our very efficient implementation of WARCA will be made publicly available.
\subsection{Datasets and Baselines}
The largest dataset we experimented with is the \textbf{Market-1501} dataset~\cite{Zheng2015} which is composed of 32,668 images of 1,501 persons captured from 6 different view points. It uses DPM~\cite{Felzenszwalb2008} detected bounding boxes as annotations. \textbf{CUHK03} dataset~\cite{Li2014}  consists of 13,164 images of 1,360 persons and it has both DPM detected and manually annotated bounding boxes. We use the manually annotated bouding boxes here.  \textbf{OpeReid} dataset~\cite{LiaoMHL14} consists of 7,413 images of 200 persons. \textbf{CUHK01} dataset~\cite{Wei2012} is composed of 3,884 images of 971 persons, with two pairs of images per person, each pair taken from a different viewpoint. Each image is of resolution 160$\times$60. The most popular and challenging person re-identification dataset is the \textbf{VIPeR}~\cite{Gray2008} dataset. It consists of 1,264 images of 632 person, with 2 images per person. The images are of resolution 128x48, captured from  horizontal viewpoints but from widely different directions. The \textbf{PRID450s} dataset~\cite{Roth2014} consists of 450 image pairs recorded from two different static surveillance cameras.  The \textbf{CAVIAR} dataset~\cite{Dong2011} consists of 1,220 images of 72 individuals from 2 cameras in a shopping mall. The number of images per person varies from 10 to 20 and image resolution also varies significantly from 141$\times$72 to 39$\times$17. The \textbf{3DPeS} dataset~\cite{Baltieri2011} has 1,011 images of 192 individuals, with 2 to 6 images per person. The dataset is captured from 8 outdoor cameras with horizontal but significantly different viewpoints. Finally the \textbf{iLIDS} dataset~\cite{Zheng2009}  contains 476 images and 119 persons, with 2 to 8 images per individual. It is captured from a horizontal view point at an airport.

We compare our method against the current state-of-the-art baselines MLAPG, rPCCA, SVMML, FRML, LFDA and KISSME. A brief overview of these methods is given in section~\ref{related_works}. rPCCA, MLAPG, SVMML, FRML are iterative methods whereas LFDA and KISSME are spectral methods on the second order statistics of the data.  Since WARCA, rPCCA  and LFDA are kernel methods we used both the $\chi^2$ kernel and the linear kernel with them to benchmark the performance.  Marginal Fisher discriminant analysis (MFA) is proven to give similar result as that of LFDA so we do not use them as the baseline. 

We did not compare against other ranking based metric learning methods such as LORETA~\cite{shalit2012}, OASIS~\cite{Chechik2010} and MLR~\cite{Mcfee2010} because all of them are linear methods. Infact we derived a kernelized OASIS but the results were not as good as ours or rPCCA. We also do not compare against LMNN and ITML because many researchers have evaluated them before~\cite{Mignon2012,Kostinger2012,Li2013} and found out that they do not perform as well as other methods considered here.

\begin{table*}[t]
  \subfloat[Rank 1 accuracy.]{
    \centering
    \resizebox{\linewidth}{!} {
\begin{tabular}{|l||cc|cc|cc|cc|cc|cc|cc|cc|cc|cc|cc|cc|cc}
\hline
Dataset & \multicolumn{2}{c|}{WARCA-$\chi^2$} & \multicolumn{2}{c|}{WARCA-L} & \multicolumn{2}{c|}{rPCCA-$\chi^2$} & \multicolumn{2}{c|}{rPCCA-L} & \multicolumn{2}{c|}{MLAPG} & \multicolumn{2}{c|}{FRML} & \multicolumn{2}{c|}{SVMML} & \multicolumn{2}{c|}{LFDA-$\chi^2$} & \multicolumn{2}{c|}{LFDA-L} & \multicolumn{2}{c|}{KISSME} \\ 
 \hline
   Market-1501& $ - $ && \textbf{45.16${\pm}$0.00}  && $-$  && $-$  && $-$  && $-$  && $-$  && $-$  && $34.65{\pm}0.00$  && $42.81{\pm}0.00$& \\ 
CUHK03&  	\textbf{78.38${\pm}$2.44} && $62.12{\pm}2.07$  && $76.74{\pm}2.06$  && $59.22{\pm}2.65$  && $44.90{\pm}1.57$  && $53.87{\pm}2.31$  && $47.89{\pm}2.59$  && $69.94{\pm}2.21$  && $46.02{\pm}1.55$  && $47.88{\pm}1.80$& \\
CUHK01&  	\textbf{58.34${\pm}$1.26} && $39.30{\pm}0.76$  && $48.55{\pm}1.12$  && $34.73{\pm}1.06$  && $22.92{\pm}0.94$  && $33.58{\pm}0.69$  && $27.96{\pm}0.86$  && $54.25{\pm}1.04$  && $33.74{\pm}0.73$  && $35.74{\pm}0.95$& \\
OpeReid&  	\textbf{57.65${\pm}$1.60} && $43.74{\pm}1.34$  && $52.89{\pm}1.78$  && $43.66{\pm}1.45$  && $40.63{\pm}1.31$  && $42.27{\pm}1.35$  && $30.63{\pm}1.51$  && $53.58{\pm}1.65$  && $42.84{\pm}1.18$  && $41.76{\pm}1.36$& \\
VIPeR&  	\textbf{37.47${\pm}$1.70} && $20.86{\pm}1.04$  && $22.25{\pm}1.91$  && $15.91{\pm}1.16$  && $19.49{\pm}2.26$  && $18.52{\pm}0.78$  && $23.28{\pm}1.53$  && $36.77{\pm}2.10$  && $20.22{\pm}1.85$  && $20.89{\pm}1.22$& \\
PRID450s&  	\textbf{24.58${\pm}$1.75} && $10.33{\pm}1.20$  && $16.35{\pm}1.30$  && $8.34{\pm}1.25$  && $2.13{\pm}0.59$  && $7.05{\pm}1.60$  && $13.08{\pm}1.63$  && $24.31{\pm}1.44$  && $3.24{\pm}0.95$  && $15.24{\pm}1.56$& \\
CAVIAR&  	\textbf{43.44${\pm}$1.82} && $39.35{\pm}1.98$  && $37.56{\pm}2.17$  && $27.26{\pm}2.15$  && $36.74{\pm}1.96$  && $35.40{\pm}2.67$  && $26.82{\pm}1.64$  && $41.29{\pm}2.25$  && $37.72{\pm}2.08$  && $31.99{\pm}2.17$& \\
3DPeS&  	\textbf{51.89${\pm}$2.27} && $43.57{\pm}2.18$  && $46.42{\pm}2.25$  && $33.12{\pm}1.58$  && $41.17{\pm}2.26$  && $39.03{\pm}1.85$  && $29.94{\pm}2.10$  && $51.44{\pm}1.40$  && $43.24{\pm}2.57$  && $37.55{\pm}1.80$& \\
iLIDS&  	\textbf{36.61${\pm}$2.40} && $31.77{\pm}2.77$  && $26.57{\pm}2.60$  && $23.07{\pm}3.07$  && $31.13{\pm}1.57$  && $25.68{\pm}2.25$  && $21.32{\pm}2.89$  && $36.23{\pm}1.89$  && $32.70{\pm}3.12$  && $28.29{\pm}3.59$& \\
\hline
\end{tabular}
 }
  \label{table_rank1}
}

\subfloat[Rank 5 accuracy.]{ 
  \centering
  \resizebox{\linewidth}{!} {
\begin{tabular}{|l||cc|cc|cc|cc|cc|cc|cc|cc|cc|cc|cc|cc|cc}
\hline
Dataset & \multicolumn{2}{c|}{WARCA-$\chi^2$} & \multicolumn{2}{c|}{WARCA-L} & \multicolumn{2}{c|}{rPCCA-$\chi^2$} & \multicolumn{2}{c|}{rPCCA-L} & \multicolumn{2}{c|}{MLAPG} & \multicolumn{2}{c|}{FRML} & \multicolumn{2}{c|}{SVMML} & \multicolumn{2}{c|}{LFDA-$\chi^2$} & \multicolumn{2}{c|}{LFDA-L} & \multicolumn{2}{c|}{KISSME} \\ 
 \hline 
  Market-1501& $ - $ && \textbf{68.23${\pm}$0.00}  && $-$  && $-$  && $-$  && $-$  && $-$  && $-$  && $52.76{\pm}0.00$  && $62.74{\pm}0.00$& \\
CUHK03&  	\textbf{94.55${\pm}$1.31} && $86.03{\pm}1.62$  && $94.50{\pm}1.29$  && $84.52{\pm}1.41$  && $71.80{\pm}1.52$  && $80.36{\pm}1.22$  && $79.97{\pm}2.08$  && $90.15{\pm}1.27$  && $65.41{\pm}1.66$  && $69.29{\pm}2.35$& \\
CUHK01&  	\textbf{79.76${\pm}$0.69} && $61.84{\pm}0.98$  && $73.29{\pm}1.32$  && $56.67{\pm}1.20$  && $48.48{\pm}1.49$  && $55.27{\pm}0.83$  && $53.11{\pm}0.78$  && $74.60{\pm}1.00$  && $49.73{\pm}0.91$  && $53.34{\pm}0.69$& \\
OpeReid&  	\textbf{80.43${\pm}$1.71} && $67.39{\pm}1.02$  && $77.95{\pm}1.82$  && $67.68{\pm}1.25$  && $61.45{\pm}1.61$  && $66.08{\pm}1.30$  && $60.32{\pm}1.31$  && $75.34{\pm}1.76$  && $59.70{\pm}1.37$  && $61.74{\pm}1.55$& \\
VIPeR&  	\textbf{70.78${\pm}$2.43} && $50.29{\pm}1.61$  && $53.82{\pm}2.32$  && $42.71{\pm}2.02$  && $46.49{\pm}2.23$  && $46.15{\pm}1.62$  && $55.28{\pm}1.99$  && $69.30{\pm}2.23$  && $45.25{\pm}1.90$  && $47.73{\pm}2.28$& \\
PRID450s&  	\textbf{55.52${\pm}$2.23} && $31.73{\pm}3.08$  && $43.82{\pm}2.18$  && $26.89{\pm}2.21$  && $11.29{\pm}1.66$  && $24.16{\pm}3.04$  && $38.38{\pm}1.77$  && $54.58{\pm}2.06$  && $12.55{\pm}1.41$  && $37.22{\pm}1.81$& \\
CAVIAR&  	\textbf{74.06${\pm}$3.13} && $68.06{\pm}2.44$  && $70.62{\pm}2.26$  && $57.44{\pm}2.48$  && $65.83{\pm}2.73$  && $66.24{\pm}3.08$  && $61.53{\pm}3.64$  && $69.12{\pm}3.02$  && $61.60{\pm}2.94$  && $61.17{\pm}3.21$& \\
3DPeS&  	\textbf{75.64${\pm}$2.80} && $68.26{\pm}1.91$  && $73.54{\pm}2.26$  && $58.34{\pm}2.31$  && $65.06{\pm}1.89$  && $65.20{\pm}2.15$  && $59.52{\pm}2.62$  && $75.36{\pm}1.91$  && $65.64{\pm}1.91$  && $60.22{\pm}2.05$& \\
iLIDS&  	\textbf{66.09${\pm}$2.31} && $59.27{\pm}3.12$  && $57.07{\pm}2.93$  && $51.55{\pm}3.59$  && $57.31{\pm}3.12$  && $53.42{\pm}2.17$  && $51.45{\pm}4.30$  && $65.20{\pm}2.68$  && $59.66{\pm}2.51$  && $54.08{\pm}3.63$& \\
\hline
\end{tabular}
  }
  \label{table_rank5}
}

\subfloat[AUC score.]{
  \centering
  \resizebox{\linewidth}{!} {
\begin{tabular}{|l||cc|cc|cc|cc|cc|cc|cc|cc|cc|cc|cc|cc|cc}
\hline
Dataset & \multicolumn{2}{c|}{WARCA-$\chi^2$} & \multicolumn{2}{c|}{WARCA-L} & \multicolumn{2}{c|}{rPCCA-$\chi^2$} & \multicolumn{2}{c|}{rPCCA-L} & \multicolumn{2}{c|}{MLAPG} & \multicolumn{2}{c|}{FRML} & \multicolumn{2}{c|}{SVMML} & \multicolumn{2}{c|}{LFDA-$\chi^2$} & \multicolumn{2}{c|}{LFDA-L} & \multicolumn{2}{c|}{KISSME} \\ 
 \hline 

      Market-1501& $ - $ && \textbf{75.41${\pm}$0.00}  && $-$  && $-$  && $-$  && $-$  && $-$  && $-$  && $60.53{\pm}0.00$  && $70.02{\pm}0.00$& \\
CUHK03&  	\textbf{93.94${\pm}$0.76} && $89.67{\pm}0.80$  && $93.92{\pm}0.81$  && $89.17{\pm}0.69$  && $82.30{\pm}1.01$  && $86.64{\pm}0.65$  && $86.64{\pm}1.07$  && $91.66{\pm}0.68$  && $74.23{\pm}1.51$  && $77.68{\pm}1.83$& \\
CUHK01&  	\textbf{84.99${\pm}$0.65} && $71.88{\pm}0.67$  && $81.00{\pm}0.88$  && $67.56{\pm}0.93$  && $62.84{\pm}1.51$  && $66.39{\pm}0.76$  && $65.73{\pm}1.07$  && $80.84{\pm}0.80$  && $58.92{\pm}1.08$  && $62.36{\pm}0.95$& \\
OpeReid&  	\textbf{86.47${\pm}$1.08} && $77.17{\pm}0.94$  && $85.25{\pm}1.16$  && $77.42{\pm}1.01$  && $72.34{\pm}1.11$  && $76.51{\pm}0.88$  && $73.88{\pm}1.04$  && $82.67{\pm}1.30$  && $68.96{\pm}1.53$  && $71.33{\pm}1.14$& \\
VIPeR&  	\textbf{81.87${\pm}$1.07} && $67.00{\pm}1.11$  && $71.30{\pm}1.50$  && $62.40{\pm}1.43$  && $64.71{\pm}1.15$  && $64.19{\pm}1.39$  && $71.04{\pm}1.63$  && $81.34{\pm}1.21$  && $62.67{\pm}1.35$  && $64.74{\pm}1.20$& \\
PRID450s&  	\textbf{72.13${\pm}$1.49} && $50.07{\pm}2.25$  && $63.10{\pm}2.16$  && $46.19{\pm}1.89$  && $30.81{\pm}2.19$  && $42.97{\pm}2.84$  && $59.54{\pm}1.25$  && $71.55{\pm}1.70$  && $28.18{\pm}1.22$  && $53.83{\pm}1.86$& \\
CAVIAR&  	\textbf{85.76${\pm}$1.48} && $83.01{\pm}1.44$  && $84.41{\pm}1.28$  && $76.57{\pm}1.29$  && $81.58{\pm}1.50$  && $81.88{\pm}1.85$  && $79.38{\pm}2.19$  && $81.94{\pm}2.32$  && $76.76{\pm}1.69$  && $78.85{\pm}1.54$& \\
3DPeS&  	\textbf{83.89${\pm}$1.53} && $78.07{\pm}1.57$  && $82.84{\pm}1.44$  && $72.27{\pm}1.96$  && $75.98{\pm}1.28$  && $76.89{\pm}1.44$  && $73.38{\pm}1.70$  && $83.49{\pm}0.95$  && $75.87{\pm}1.49$  && $72.22{\pm}1.31$& \\
iLIDS&  	\textbf{79.04${\pm}$1.60} && $73.42{\pm}1.96$  && $74.10{\pm}2.04$  && $69.60{\pm}2.44$  && $72.45{\pm}1.99$  && $71.26{\pm}1.55$  && $70.25{\pm}2.09$  && $78.98{\pm}1.43$  && $74.26{\pm}2.02$  && $70.33{\pm}2.90$& \\
\hline
\end{tabular}
  }
  \label{table_auc}
}
\caption{Table showing the rank 1, rank 5 and AUC performance measure of our method WARCA against other state-of-the-art methods. Bold fields indicate best performing methods. The dashes indicate computation that could not be run  in a realistic setting on Market-1501.}
\end{table*}

\begin{figure*}[!t]
\centering
\centering
\resizebox{\linewidth}{!}{
\begin{tabular}{ccc}
\includegraphics[width=\textwidth]{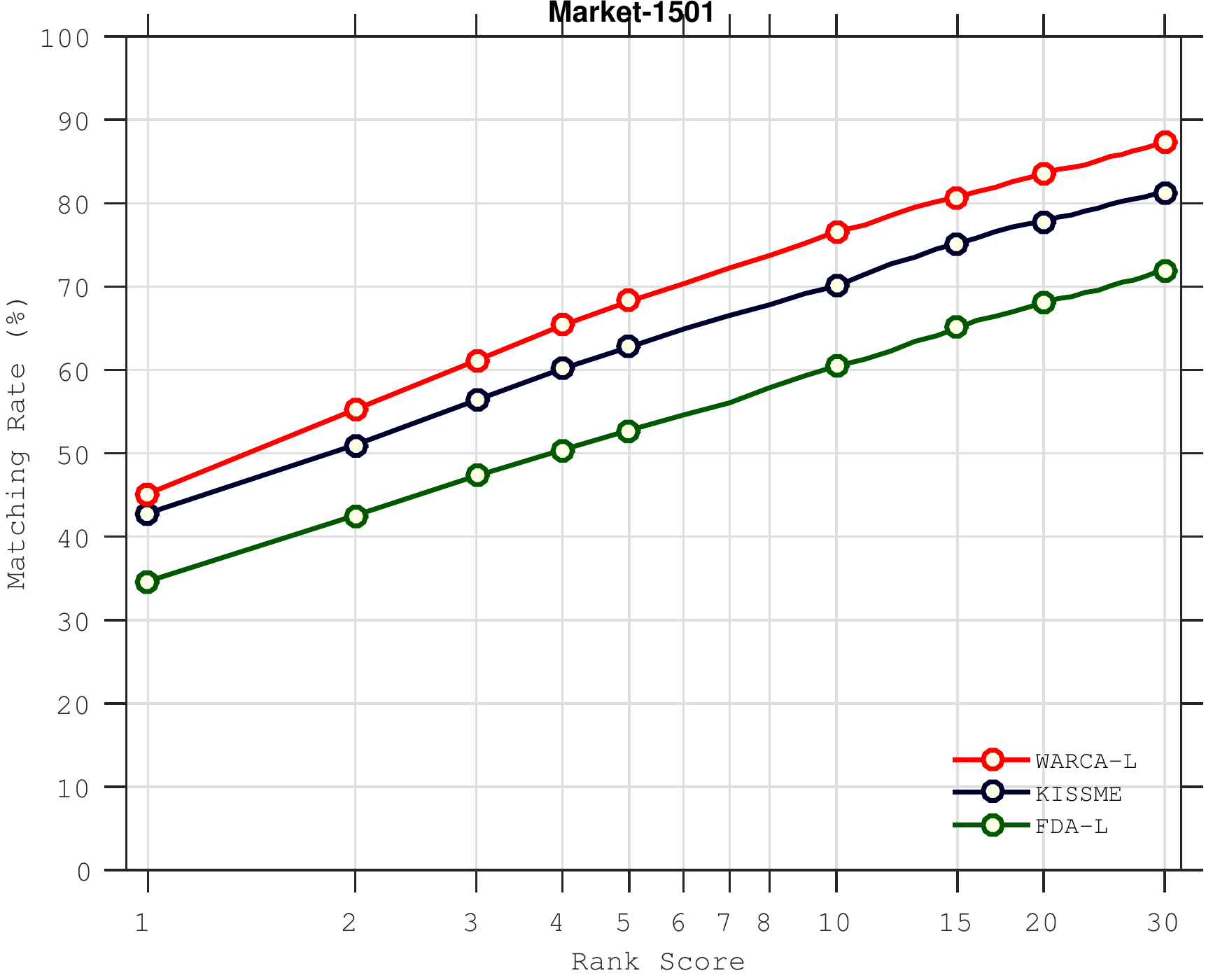} &
\includegraphics[width=\textwidth]{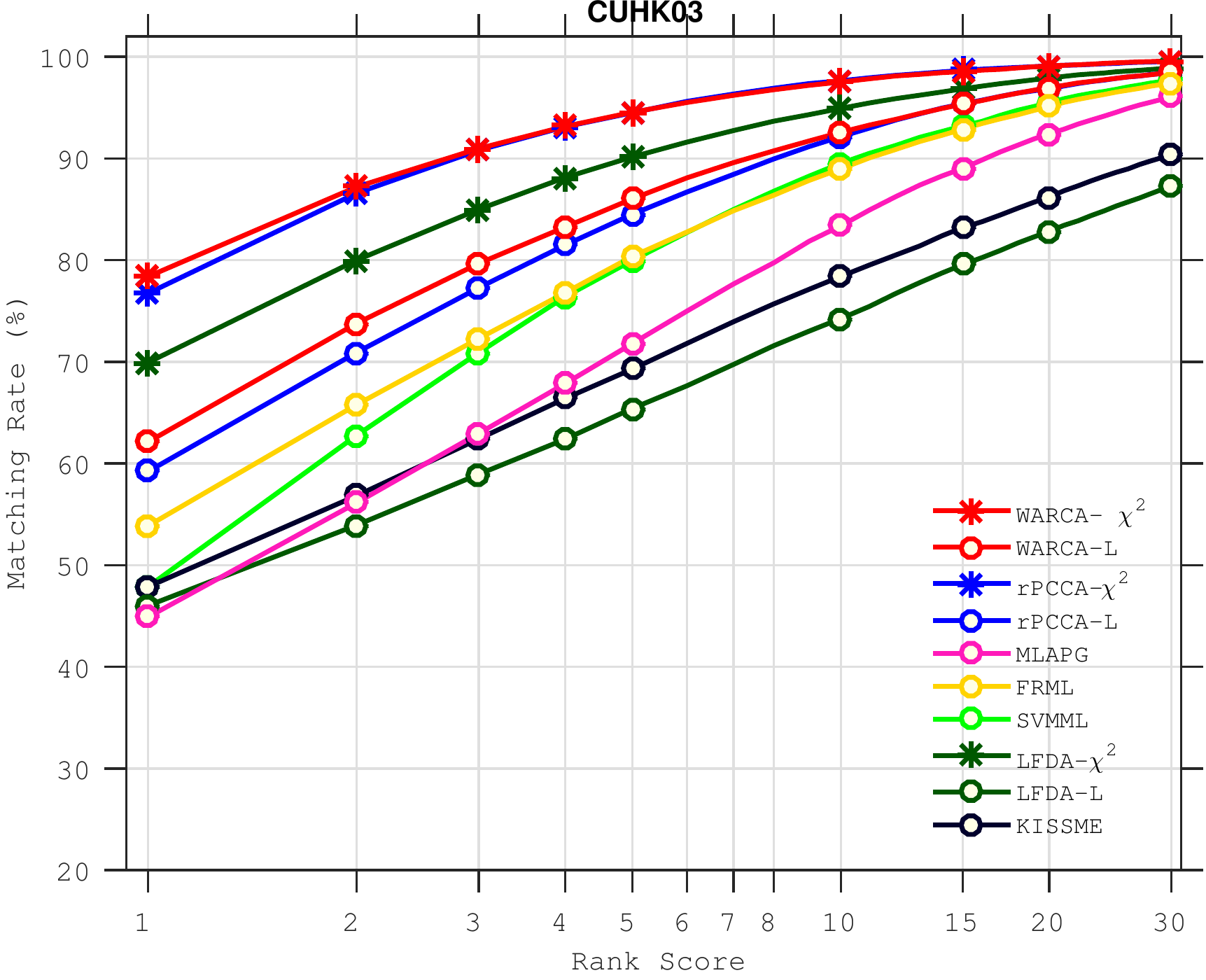} &
\includegraphics[width=\textwidth]{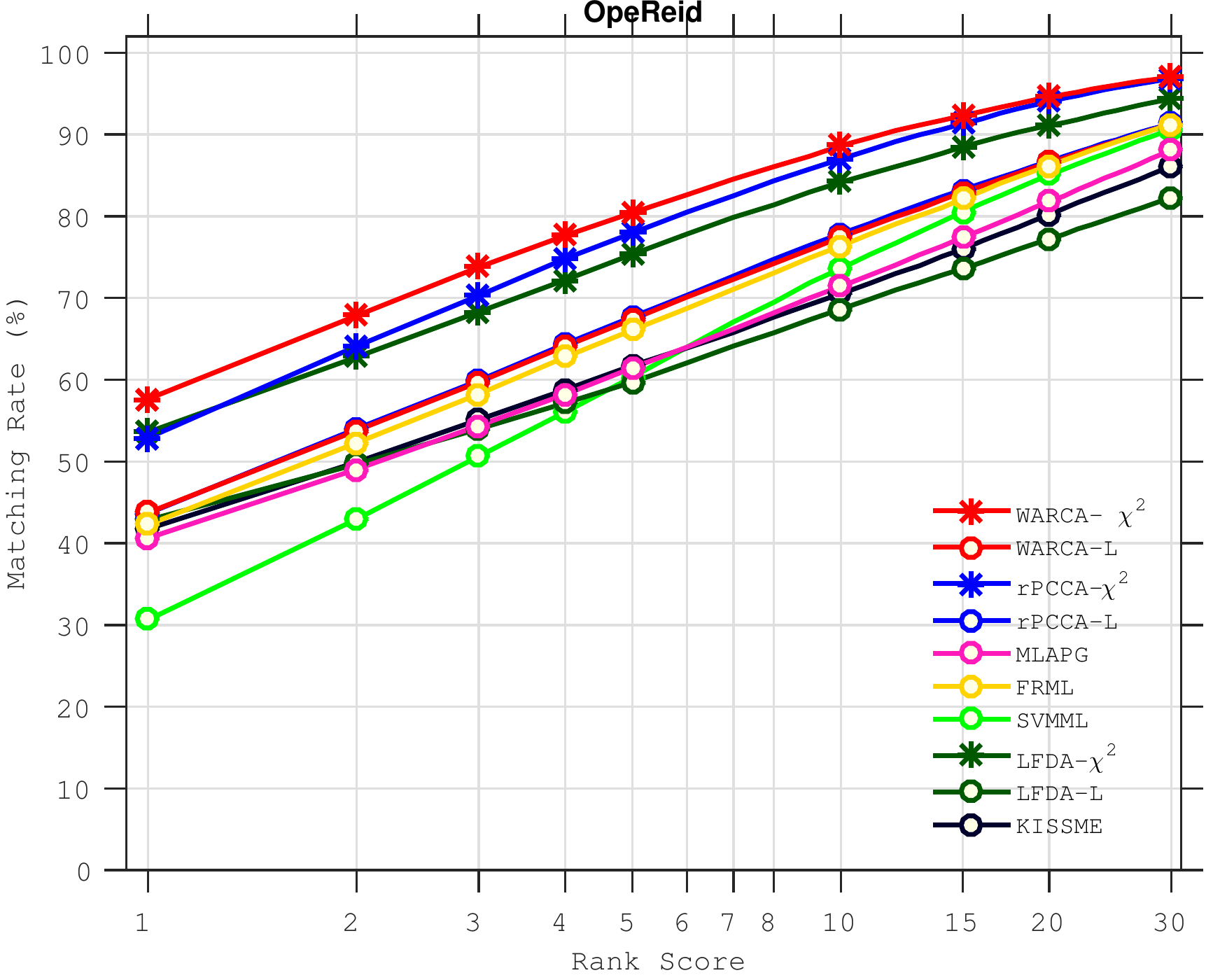} \\ \\
\includegraphics[width=\textwidth]{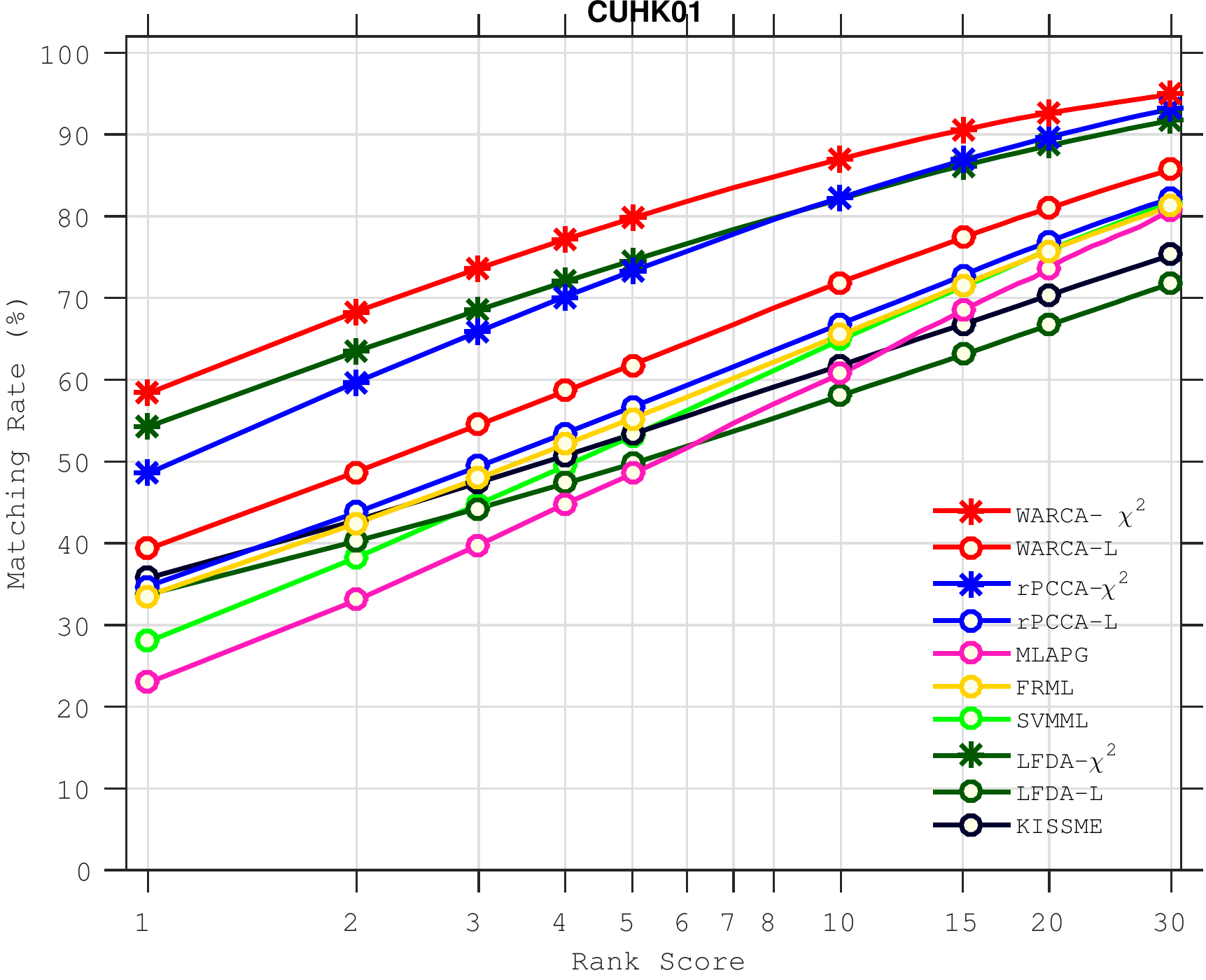} &
\includegraphics[width=\textwidth]{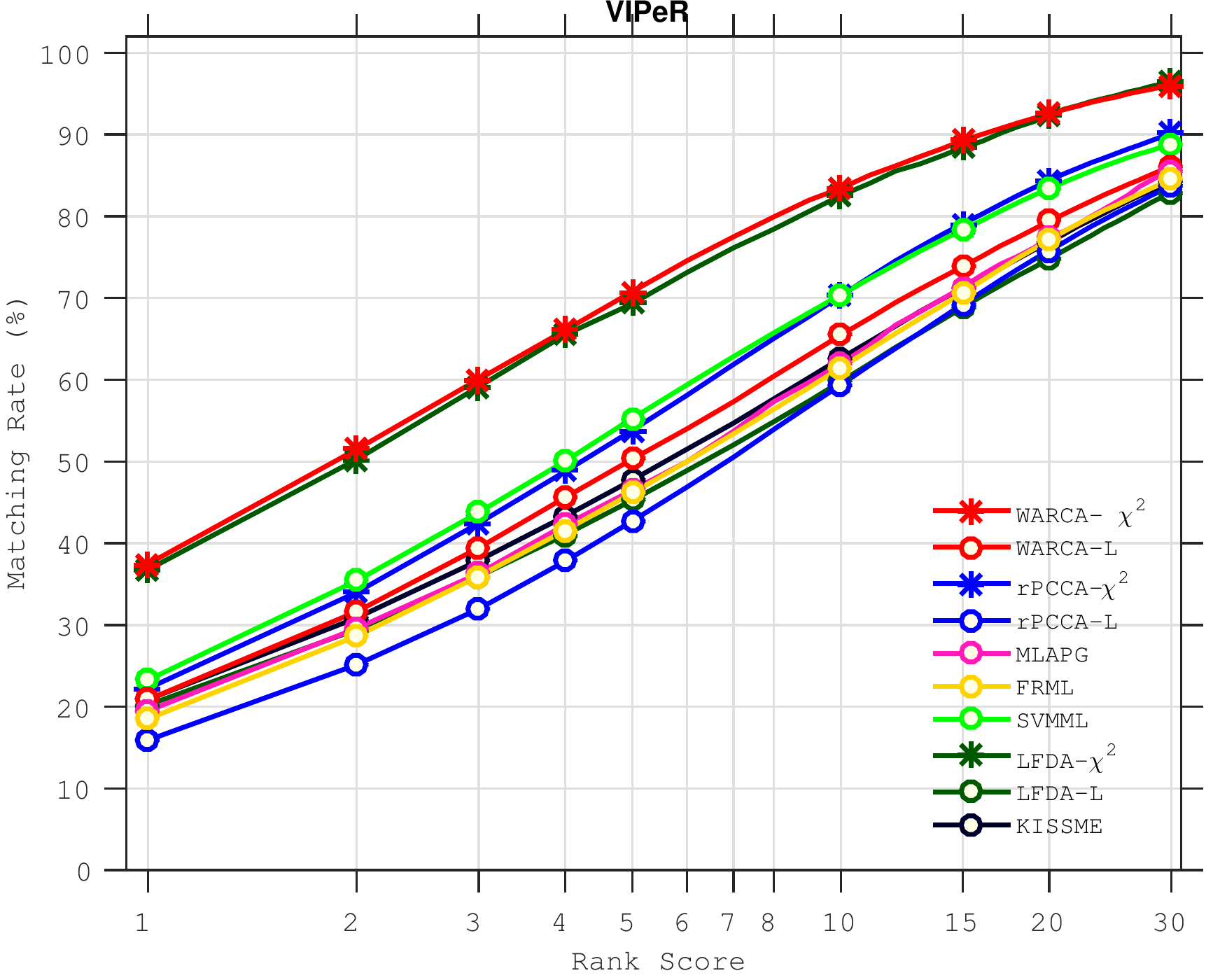}  &
\includegraphics[width=\textwidth]{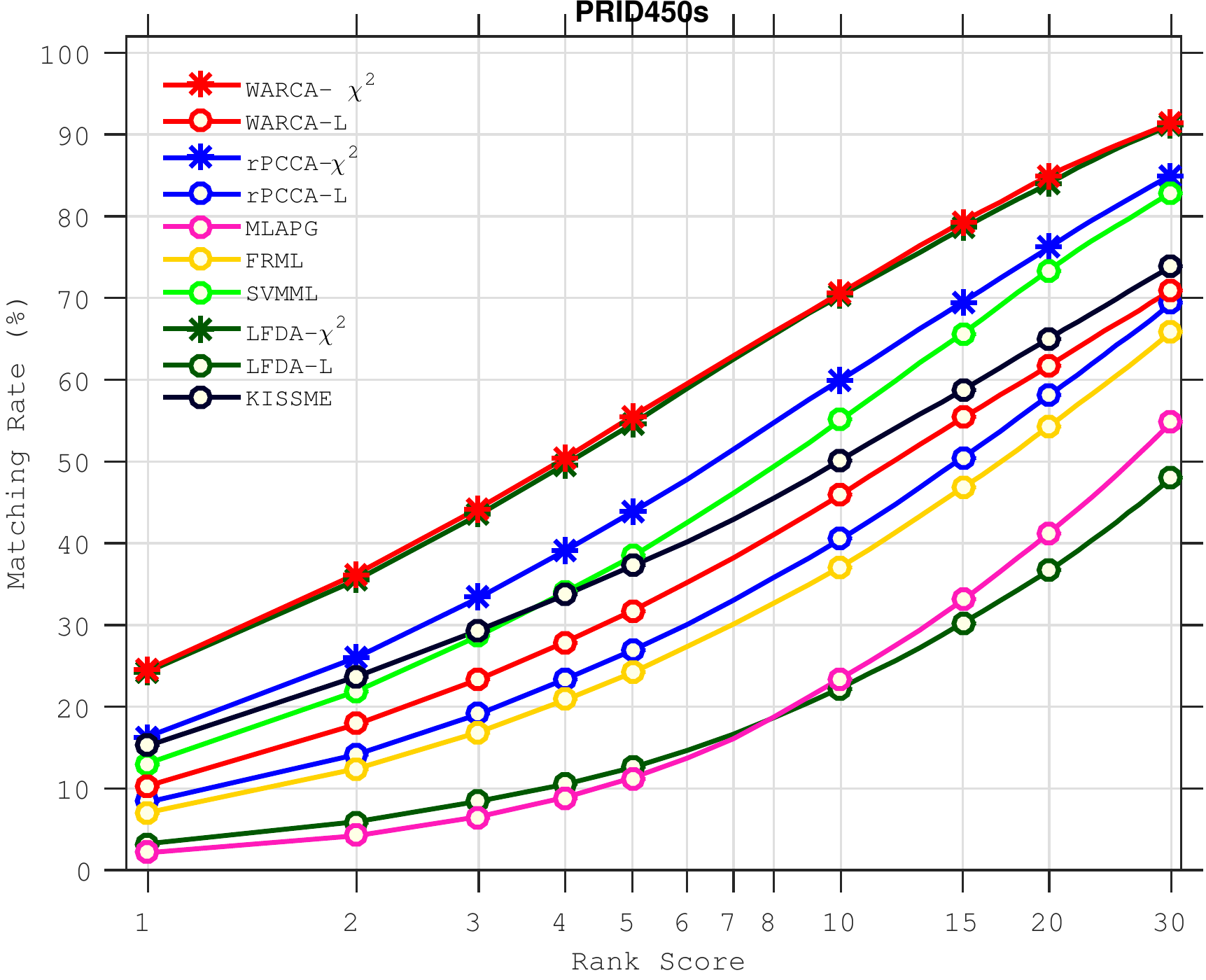} \\ \\
\includegraphics[width=\textwidth]{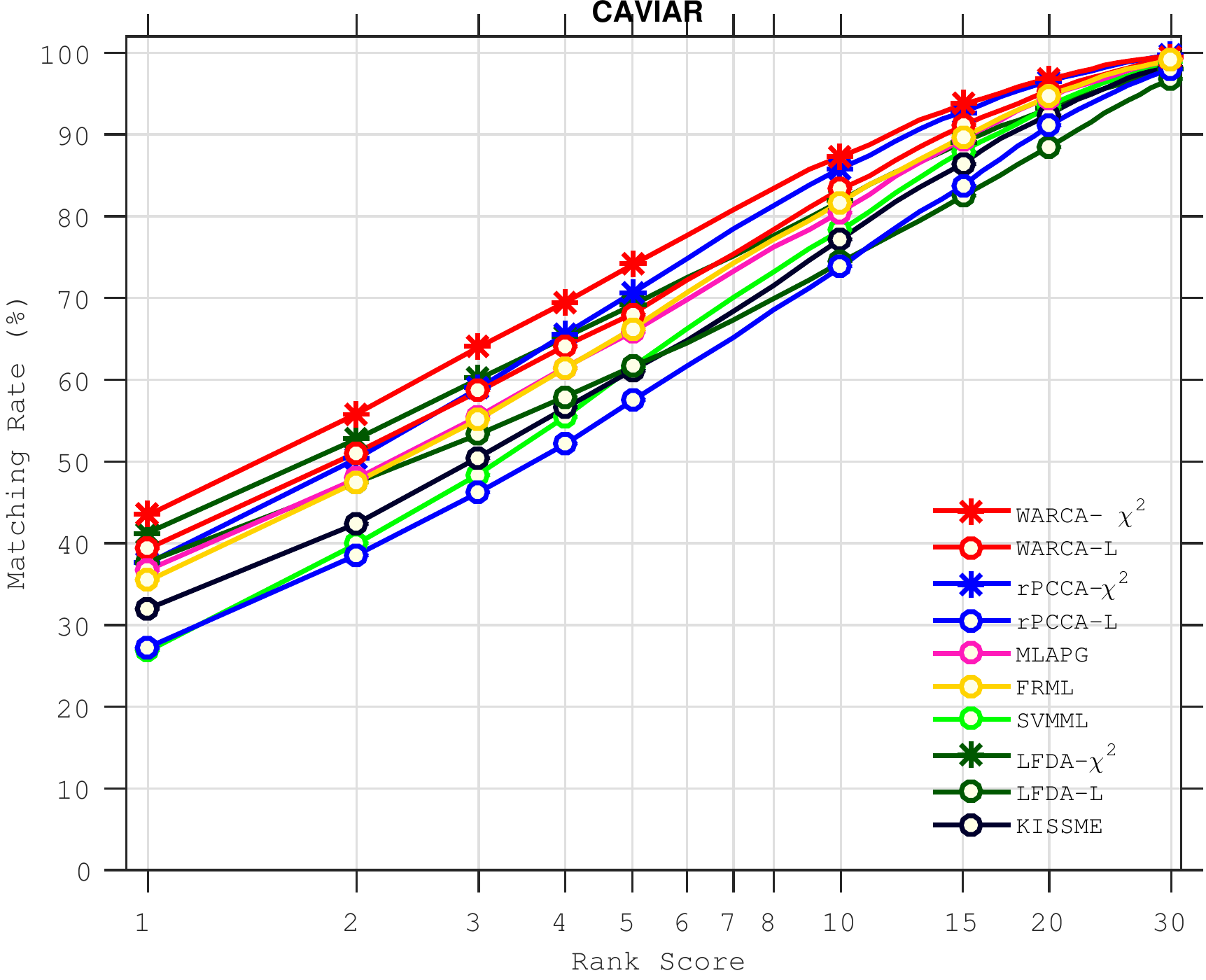} &
\includegraphics[width=\textwidth]{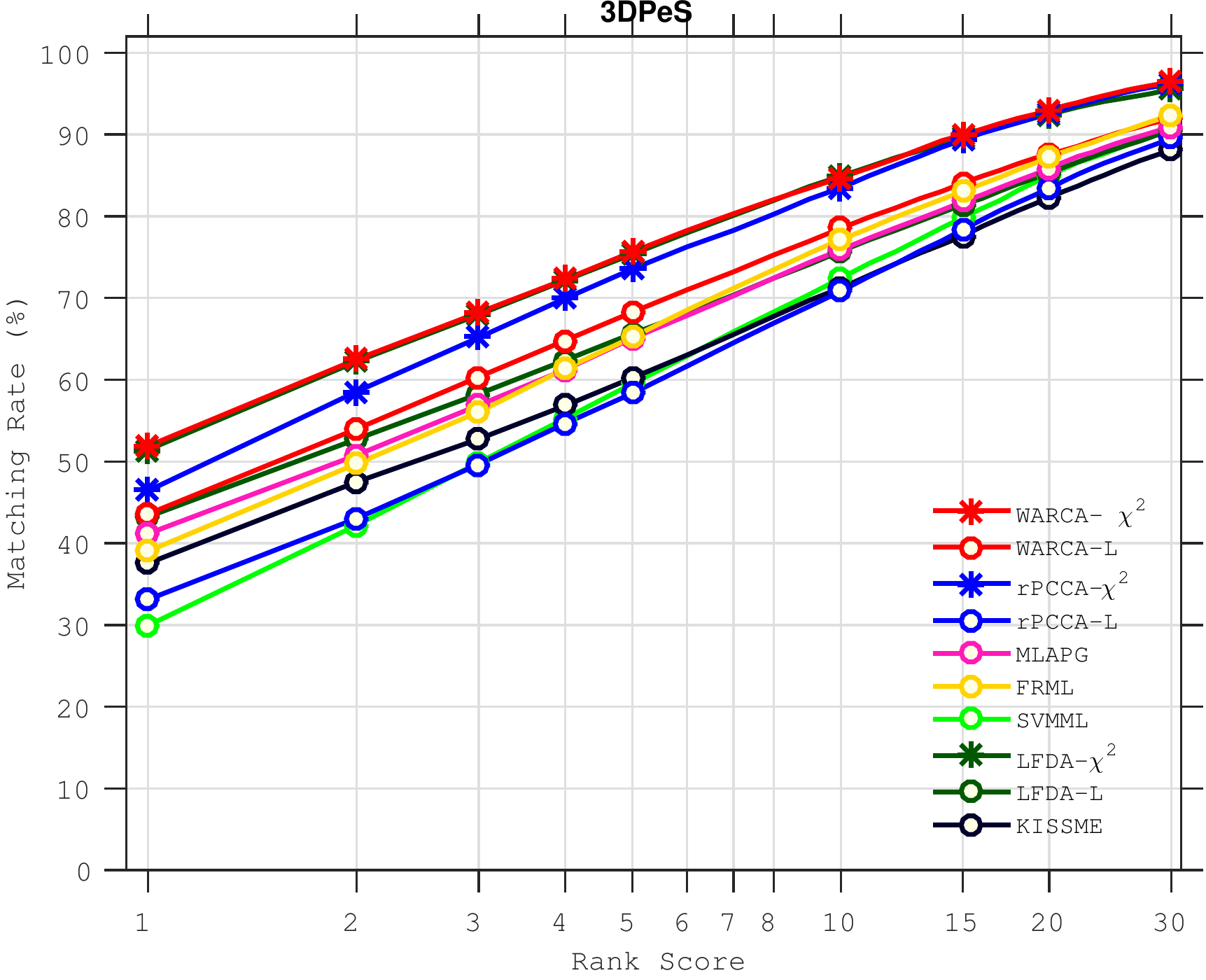} &
\includegraphics[width=\textwidth]{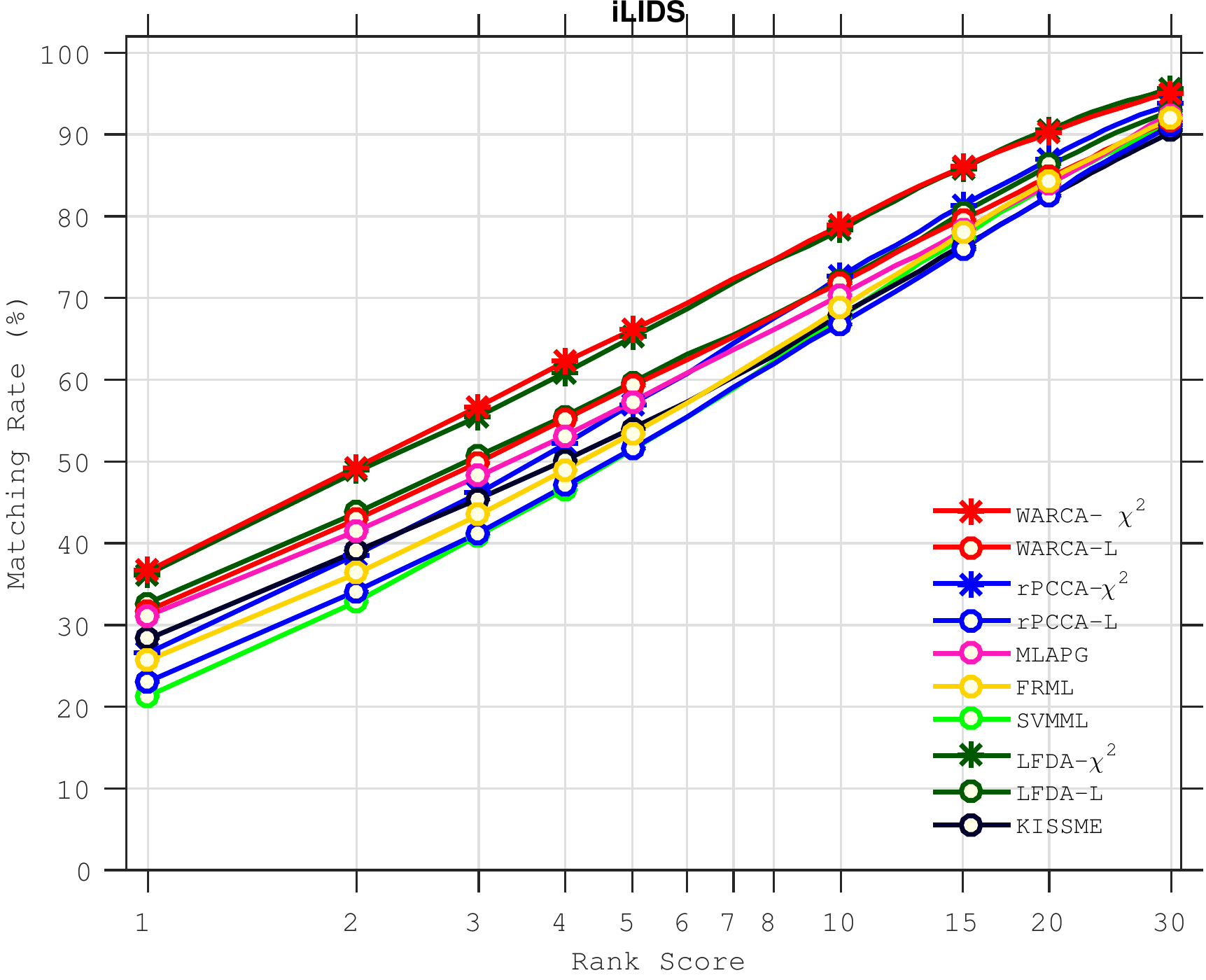}
\end{tabular}
}
\caption{CMC curves comparing WARCA against state-of-the-art methods on nine re-identification datasets}
\label{fig_cmcfda}
\end{figure*}

\subsection{Technical Details}
For the Market-1501 dataset we used the experimental protocol and features described in~\cite{Zheng2015}. We used their baseline code and features. As Market-1501 is quite large for kernel methods we do not evaluate them. We also do not evaluate the linear methods such as Linear rPCCA and SVMML because their optimization algorithms were found to be very slow.

All other evaluations where carried out in the single-shot experiment setting~\cite{Gong2014} and our experimental settings are very similar to the one adopted by Xiong\etal~\cite{Xiong2014}. Except for Market-1501, we randomly divided all the other datasets  into two subsets such that there are $p$ individuals in the test set. We created 10 such random splits. In each partition one image of each person was randomly selected as a probe image, and the rest of the images were used as gallery images and this was repeated 10 times. The position of the correct match was processed to generate the CMC curve. We followed the standard train-validation-test splits for all the other datasets and $P$ was chosen to be 100, 119, 486, 316, 225, 36, 95 and 60 for CUHK03, OpeReid, CUHK01, VIPeR, PRID450s, CAVIAR, 3DPeS and iLIDS respectively.

We used the same set of features for all the datasets except for the Market-1501 and all the features are essentially histogram based. First all the datasets were re-scaled to 128$\times$48 resolution and then 16 bin color histograms on RGB, YUV, and HSV channels, as well as texture histogram based on Local Binary Patterns (LBP) were extracted on 6 non-overlapping horizontal patches. All the histograms are normalized per patch to have unit $L_1$ norm and  concatenated into a single vector of dimension 2,580~\cite{Mignon2012,Xiong2014}.

The source codes for LFDA, KISSME and SVMML are available from their respective authors website, and we used those to reproduce the baseline results~\cite{Xiong2014}. The code for PCCA is not released publicly. A version from  Xiong\etal~\cite{Xiong2014} is available publicly but the memory footprint of that implementation is very high making it impossible to use with large datasets (e.g. it requires 17GB of RAM to run on the CAVIAR dataset). Therefore to reproduce the results in~\cite{Xiong2014} we wrote our own implementation, which uses 30 times less memory and can scale to much larger datasets. We also ran sanity checks to make sure that it behaves the same as that of the baseline code. All the implementations were done in Matlab with mex functions for the acceleration of the critical components. 

In order to fairly evaluate the algorithms, we set the dimensionality of the projected space to be same for WARCA, rPCCA and LFDA. For the Market-1501 dataset the dimensionality used is 200 and for VIPeR it is 100 and  all the other datasets it is 40. We choose the regularization parameter and the learning rate  through cross-validation across the data splits using grid search in  $(\lambda, \eta) \in \{ 10^{-8}, \dots, 1 \}
\times \{ 10^{-3}, \dots, 1 \}$. Margin $\gamma$ is fixed to $1$. Since the size of the parameter matrix scales in $O(D^2)$ for SVMML and KISSME we first reduced the dimension of the original features using PCA keeping 95$\%$ of the original variance and then applied these algorithms. In our tables and figures WARCA$-\chi^2$, WARCA-L, rPCCA$-\chi^2$,  rPCCA-L,  LFDA$-\chi^2$ and LFDA-L denote WARCA with $\chi^2$ kernel, WARCA with linear kernel, rPCCA with $\chi^2$ kernel, rPCCA with linear kernel,  and LFDA with $\chi^2$ kernel, LFDA with linear kernel respectively.

For all experiments with WARCA we used harmonic weighting for the rank weighting function of Equation~\ref{equation_wr}. That is weighting of the form ${{L}}(M) = \sum_{m=1}^{M} \frac{1}{m}$. We also tried uniform weighting which gave poor results compared to the harmonic weighting for a given computational budget. For all the datasets we used a mini-batch size of 512 in the SGD algorithm and we ran the SGD for 2000 iterations (A parameter update using the mini-batch is considered as 1 iteration).

Tables~\ref{table_rank1} and~\ref{table_rank5} summarize respectively the rank-1 and rank-5 performance of all the methods, and Table~\ref{table_auc} summarizes the Area Under the Curve (AUC) performance score. Figure~\ref{fig_cmcfda} reports the CMC curves comparing WARCA against the  baselines on all the nine datasets. The circle and the star markers denote linear and kernel methods respectively.

WARCA improves over all other methods on all the datasets. On VIPeR, 3DPeS, PRID450s and iLIDS  datasets LFDA come very close to  the performance of WARCA. The reason for this is that these datasets are too small and consequently simple methods such as LFDA which exploits strong prior assumptions on the data distribution work nearly as well as WARCA.

\subsection{Comparison against State-of-the-art}
\begin{table}[!]
\centering
\resizebox{\columnwidth}{!}{%
\begin{tabular}{|c|cccc|cccc|cccc|cccc|}
\hline
\multirow{2}{*}{Dataset} &
\multicolumn{4}{c|}{WARCA(Ours)}  & \multicolumn{4}{c|}{MLAPG~\cite{LiaoMLAPG2015}} & \multicolumn{4}{c|}{MLPOLY~\cite{Chen2015}} & \multicolumn{4}{c|}{IDEEP~\cite{Ahmed2015}} \\
\cline{2-17}
  & rank=1 & rank=5 & rank=10 & rank=20 & rank=1 & rank=5 & rank=10 & rank=20 & rank=1 & rank=5 & rank=10 & rank=20 & rank=1 & rank=5 & rank=10 & rank=20 \\
\hline
VIPeR & 40.22 & 68.16 & 80.70 & 91.14  & \textbf{40.73} & \textbf{69.94} & \textbf{82.34} & \textbf{92.37} & 36.80 & 70.40 & 83.70 & 91.70 & 34.81 & 63.61 & 75.63 & 84.49  \\
CUHK01 & \textbf{65.64} & 85.34 & 90.48 & \textbf{95.04} & 64.24 & \textbf{85.41} & \textbf{90.84} & 94.92 & - & - & - & -& 47.53 & 71.60 & 80.25 & 87.45  \\
CUHK03 & \textbf{78.38} & \textbf{94.5} & \textbf{97.52} & \textbf{99.11} & 57.96 & 87.09 & 94.74 & 98.00 & - & - & - & -& 54.74 & 86.50 & 94.02 & 97.02  \\
\hline
\end{tabular}
}
\caption{Comparison of WARCA against state-of-the-art results for person re-identification. \label{table_add}}
\end{table}

We also compare against the state-of-the-art results reported using recent algorithms such as MLAPG on LOMO features~\cite{LiaoMLAPG2015}, MLPOLY~\cite{Chen2015} and IDEEP~\cite{Ahmed2015} on VIPeR, CUHK01 and CUHK03 datasets. The reason for not including these comparisons in the main results is because apart from MLAPG  the code for other methods is not available, or the features are different which makes a fair comparison difficult. Our goal is to evaluate experimentally that, given a set of features, which is the best off-the-shelf metric learning algorithm for re-identification.

In this set of experiments we used the state-of-the-art LOMO features~\cite{LiaoXQDA2015} with WARCA for VIPeR and CUHK01 datasets. The results are summarized in the Table~\ref{table_add}. We improve the rank1 performance by $21\%$ on CUHK03 by $1.40\%$ on CUHK01 dataset.

\subsection{Analysis of the AON regularizer}\label{sec:analys-aon-regul}

\begin{figure*}
\hspace*{-0.1cm}
\subfloat[]{\label{ref_cond}\includegraphics[scale=0.17]{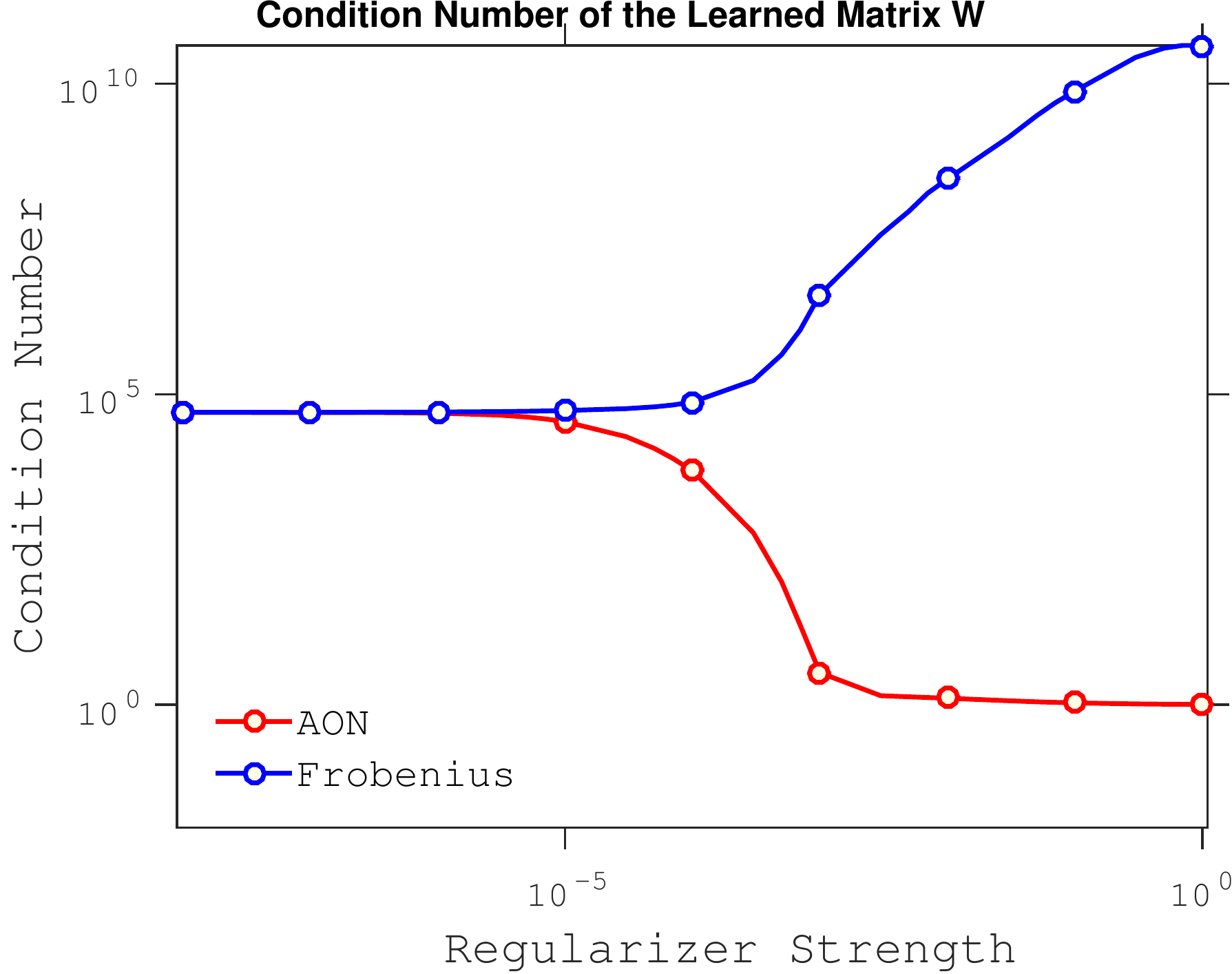}}
\subfloat[]{\label{ref_lambda}\includegraphics[scale=0.17]{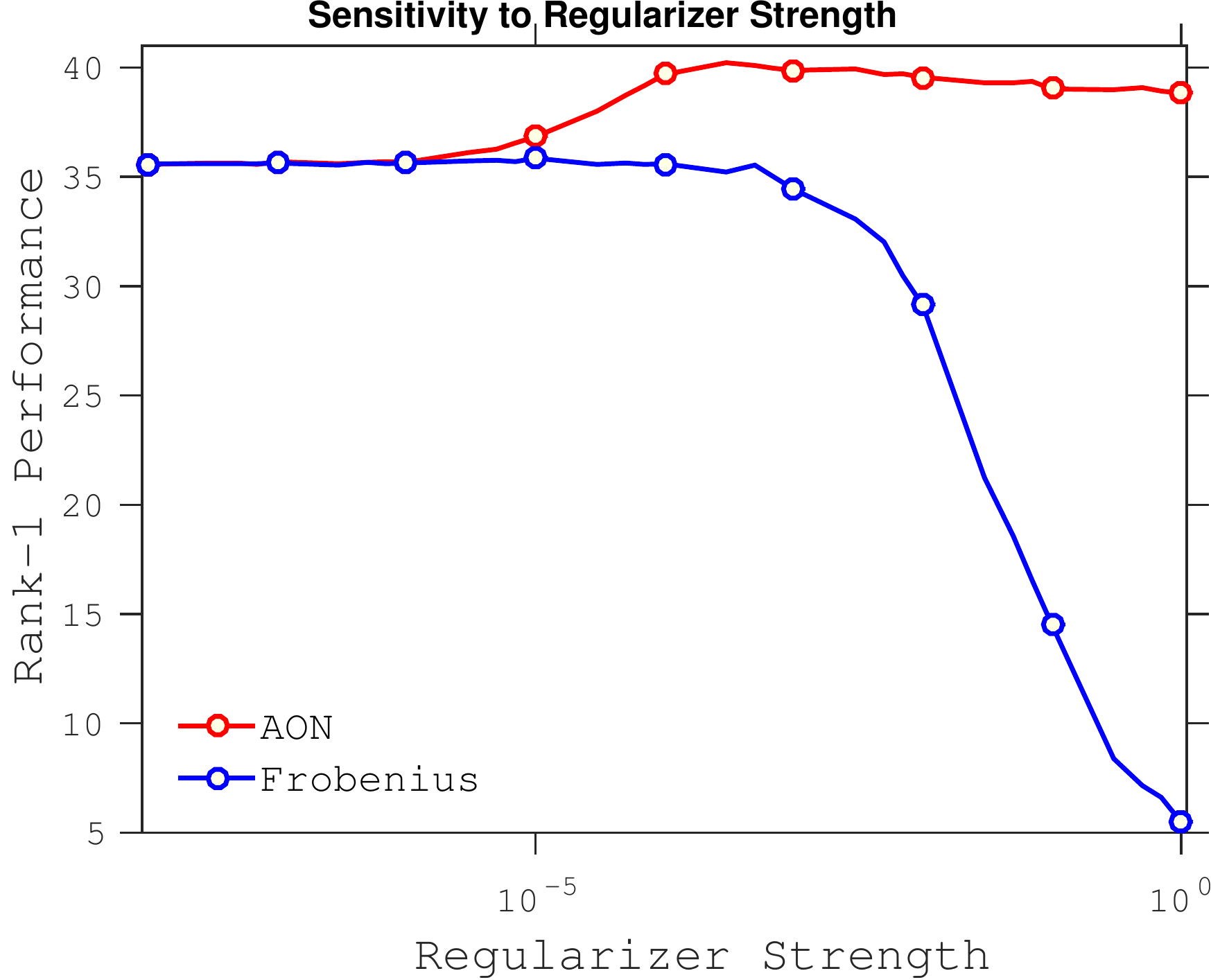}}
\subfloat[]{\label{ref_eta}\includegraphics[scale=0.17]{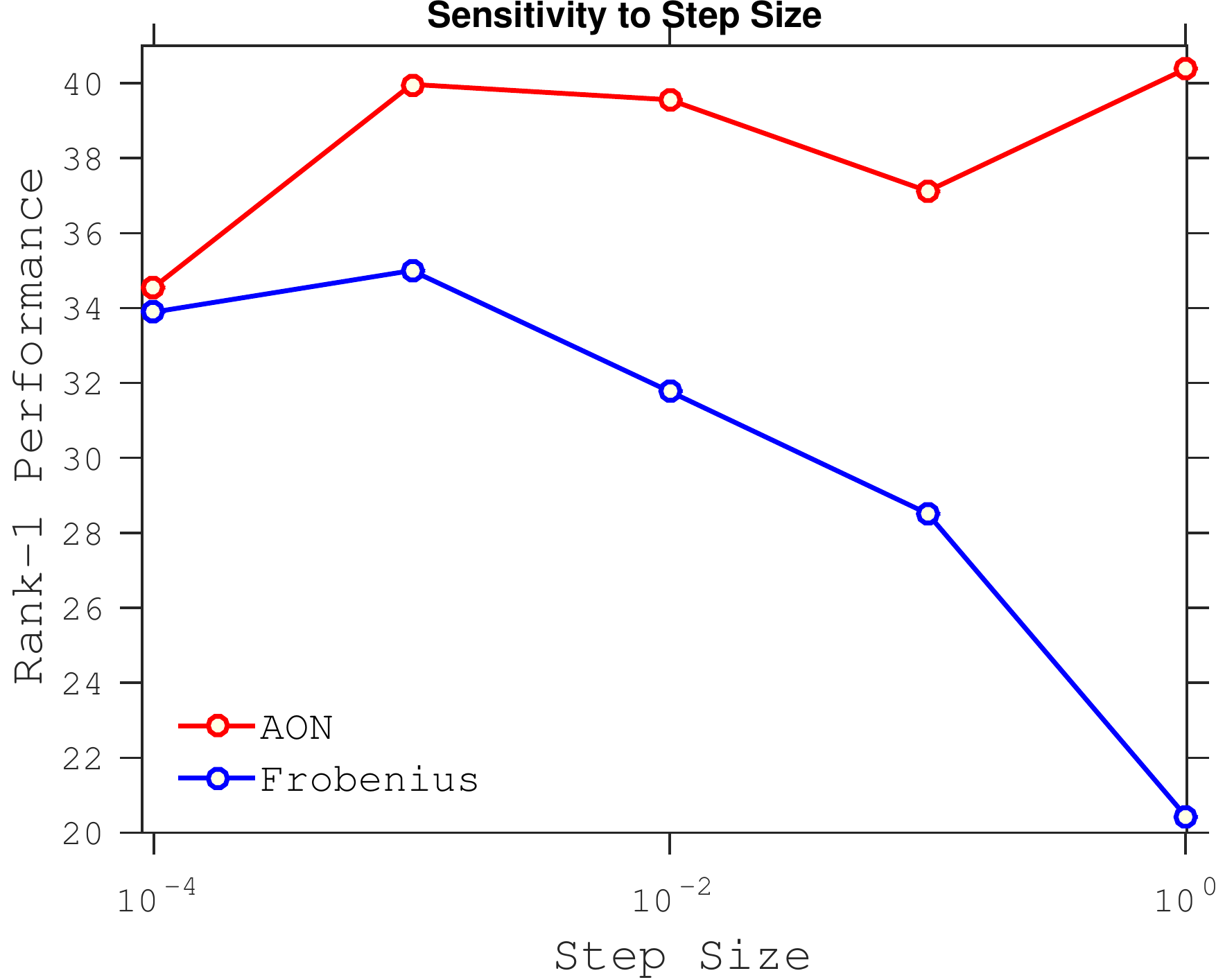}}
\subfloat[]{\label{ref_cmc}\includegraphics[scale=0.17]{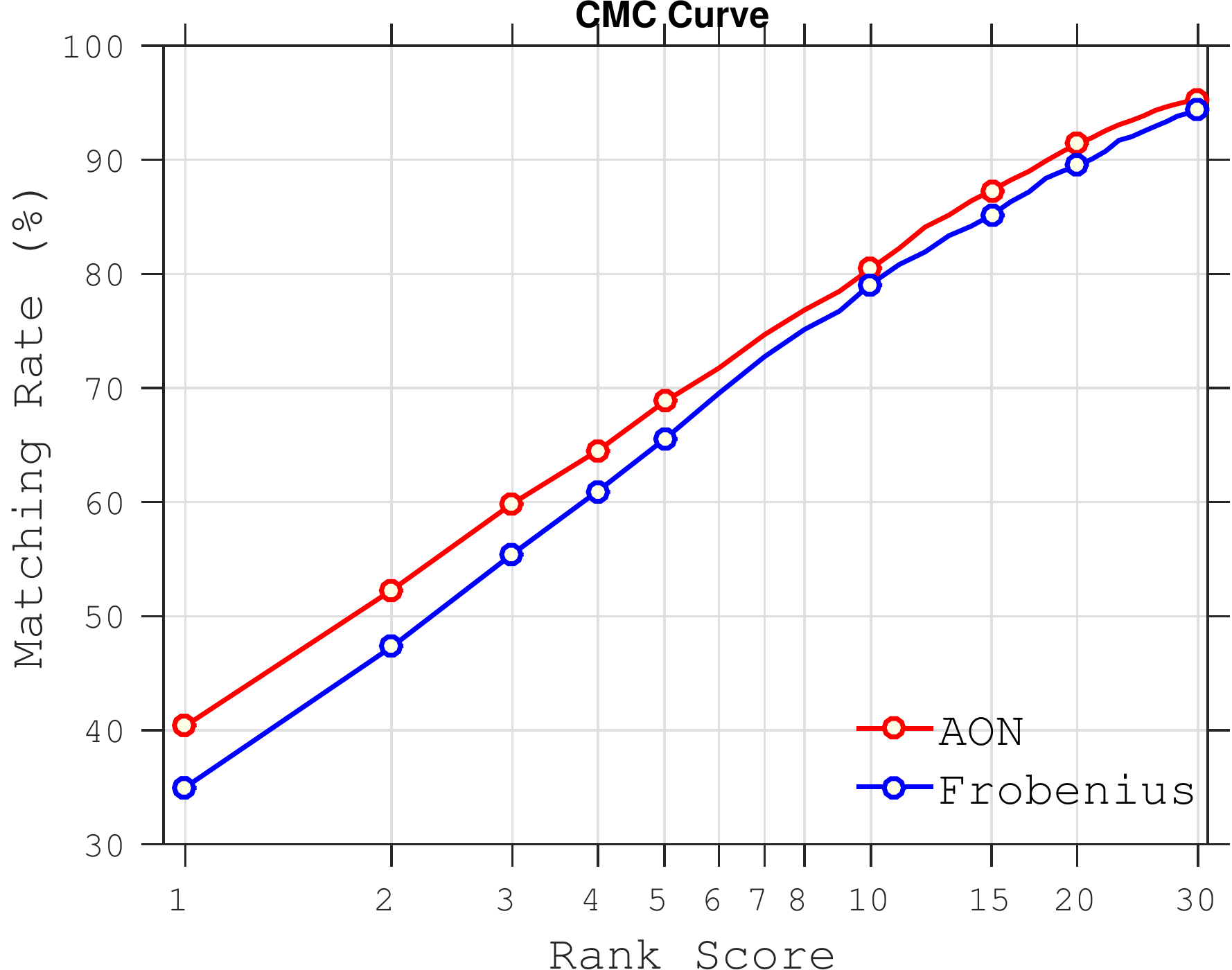}}
\caption{
Comparison of the Approximate OrthoNormal (AON) regularizer we use in our algorithm to the standard Frobenius norm ($L_2$) regularizer.
Graph \protect\subref{ref_cond} shows the condition number (ratio between the two
extreme eigenvalues of the learned mapping) vs. the weight $\lambda$
of the regularization term. As expected, the AON regularizer pushes
this value to one, as it eventually forces the learning to chose an
orthonormal transformation, while the Frobenius regularizer eventually kills
the smallest eigenvalues to zero, making the ratio extremely large. Graph
\protect\subref{ref_lambda} shows the Rank-1 performance vs. the regularizer
weight $\lambda$, graph \protect\subref{ref_eta} the Rank-1 performance
vs. the SGD step size $\eta$, and finally graph \protect\subref{ref_cmc} CMC
curve with the two regularizers.
\label{ref_reg}}
\end{figure*}

Here we present an empirical analysis of the AON regularizer against the standard Frobenius norm regularizer. We used the VIPeR dataset with LOMO features for all these experiments.
With very low regularization strength AON and Frobenius behaves the same. As the regularization strength increases Frobenous norm results in rank deficient mappings (Figure~\ref{ref_cond}), which is less discriminant and performs poorly on the test set (Figure~\ref{ref_lambda}). On the contrary, the AON regularizer pushes towards orthonormal mappings and results in an embedding well conditioned, which generalizes well to the test set. It is also worth noting that training with the AON regularizer is robust over wide range of regularization parameter, which is not the case the Frobenius norm.

Finally, the AON regularizer was found to be very robust to the choice of the SGD step size $\eta$ (Figure~\ref{ref_eta}) which is a crucial parameter in large-scale learning. A similar behaviour was observed by Lim\etal~\cite{Lim2014} with their orthonormal Riemannian gradient update step in the SGD but it is computationally very expensive and cannot be used combined with modern SGD algorithms such as Adam~\cite{Kingma2014}, and Nesterov's momentum~\cite{Sutskever2013}.

\subsection{Analysis of the Training Time}

\begin{figure*}
\centering
\centering
\resizebox{\linewidth}{!}{
\begin{tabular}{ccc}
\includegraphics[width=\textwidth]{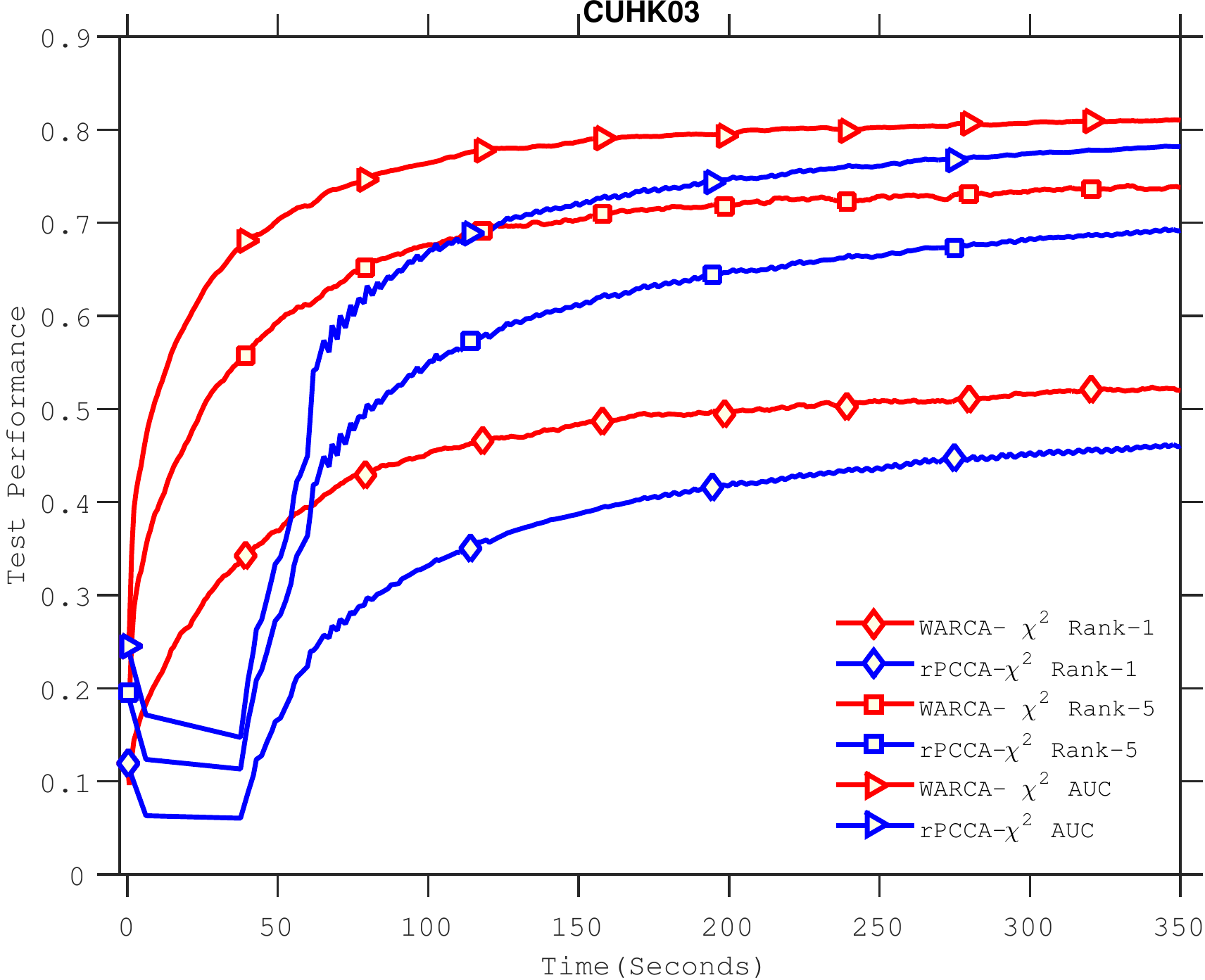} &
\includegraphics[width=\textwidth]{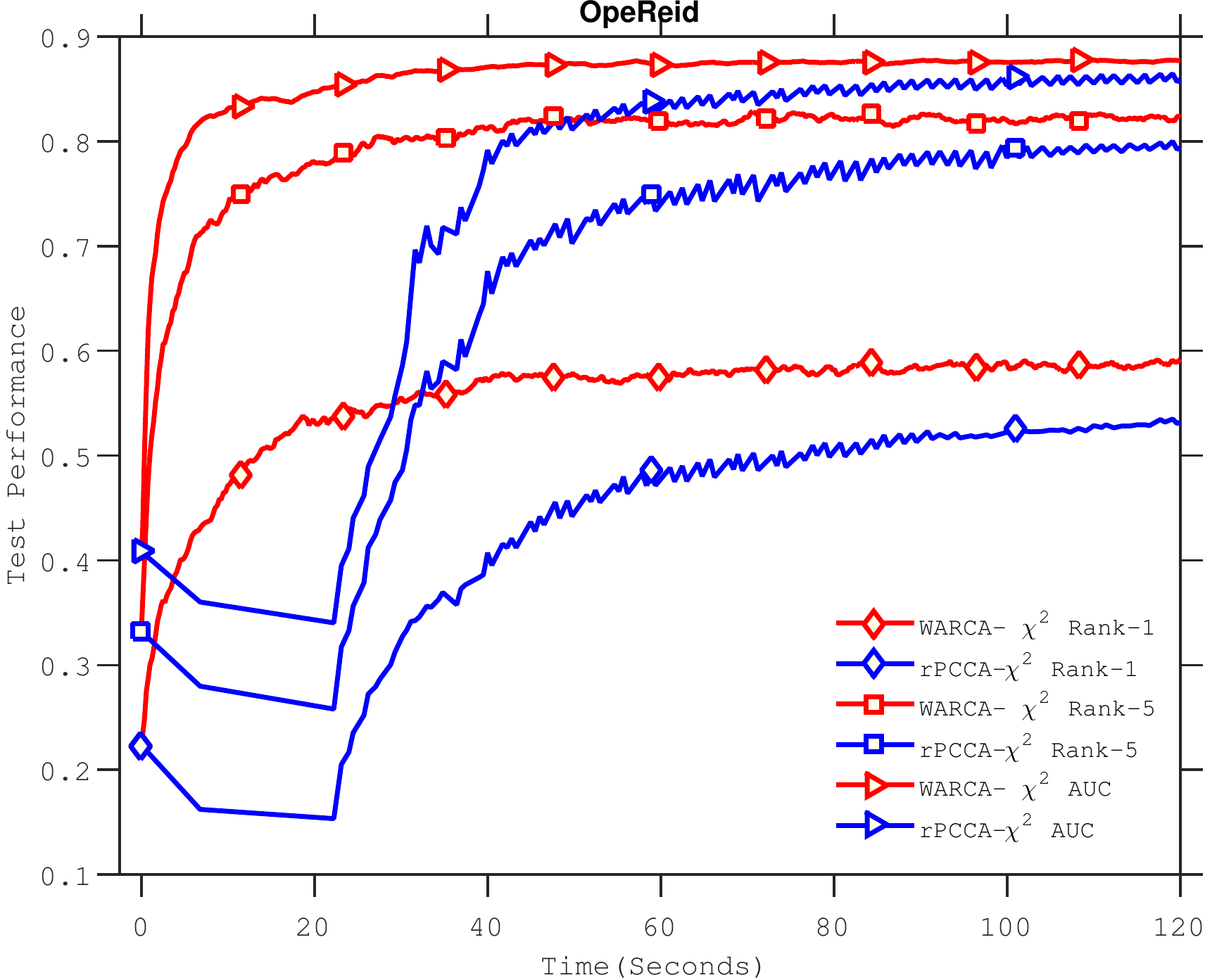}
\includegraphics[width=\textwidth]{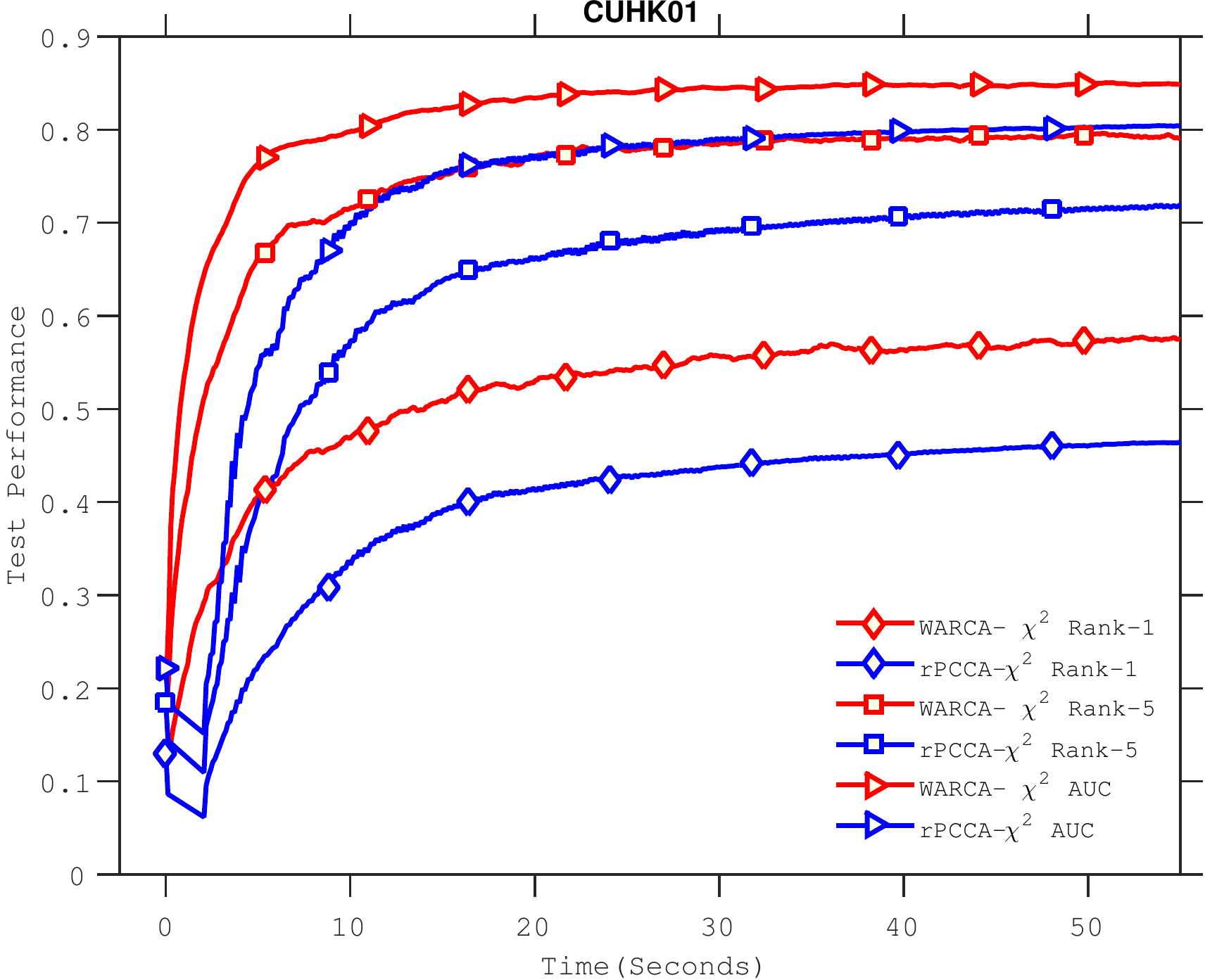}
\end{tabular}
}
\caption{WARCA performs significantly better than the state-of-the-art rPCCA on large datasets for a given training time budget.}
\label{fig_WARCAvsrPCCA}
\end{figure*}

Figure~\ref{fig_WARCAvsrPCCA} illustrates how the test set performance of WARCA and rPCCA increase as a function of training time on 3 datasets. We implemented both the  algorithms entirely in C++ with BLAS and OpenMP to have a fair comparison on their running times. In this set of experiments  we used number of test identites to be 730 for CUHK03 dataset to have a quick evaluation. Other datasets follow the same experimental protocol described above.   Please note that we do not include spectral methods in this plot because its solutions are found analytically. Linear spectral methods are very fast for low dimensional problems but the training time scales quadratically in the data dimension. In case of kernel spectral methods the training time scales quadratically in the number of data points. We also do not include iterative methods, MLAPG and SVMML because they proved to be very slow and not giving good performance.


\section{Conclusion and Future work}

We have proposed a simple and scalable approach to metric learning that combines a new and simple regularizer to a proxy for a weighted sum of the precision at different ranks. The later can be used for any weighting of the precision-at-$k$ metrics. Experimental results show that it outperforms state-of-the-art methods on standard person re-identification datasets,  and that contrary to most of the current state-of-the-art methods, it allows for large-scale learning.


The simplicity and efficiency of WARCA call for several future
research directions. The first one is to investigate different forms of
the regularizer, in particular using the LogDet
divergence~\cite{James1961} instead of the $L_2$. Such a form is
justifiable under a Gaussian assumption on the data distributions, and
it has been used successfully for metric learning
before~\cite{Davis2007}. The second is to extend WARCA to a more
general class of models, in particular non-parametric ones such as
forests of decision trees or (deep) multi-layer perceptrons.

From a more theoretical perspective, we are interested also in looking
at the relations between the behavior of the learning with the
orthonormal regularizer, and the recent residual
networks~\cite{DBLP:journals/corr/HeZRS15}. In both case, strong
regularization pushes toward full-rank mappings instead of null
transformations, as standard $L_2$ penalty does, which appears to be a
very reasonable behavior to expect in general.

\bibliographystyle{splncs}
\bibliography{reident}

\checknbdrafts

\end{document}